\documentclass{article} % For LaTeX2e
\usepackage{iclr2027_arxiv,times}

\usepackage{amsmath,amsfonts,bm}

\def\eqref#1{equation~\ref{#1}}
\def\1{\bm{1}}

\DeclareMathAlphabet{\mathsfit}{\encodingdefault}{\sfdefault}{m}{sl}
\SetMathAlphabet{\mathsfit}{bold}{\encodingdefault}{\sfdefault}{bx}{n}

\usepackage{hyperref}
\usepackage{url}
\usepackage{booktabs}       % professional-quality tables
\usepackage{amsfonts}       % blackboard math symbols
\usepackage{nicefrac}       % compact symbols for 1/2, etc.
\usepackage{microtype}      % microtypography
\usepackage{xcolor}         % colors
\usepackage{xspace}
\usepackage{enumitem}
\usepackage{amsmath,amsthm,amssymb,amsfonts,dsfont,pifont,bm,bbm,mathrsfs,mathtools,nicefrac,extarrows,relsize}
\usepackage{algorithm,algpseudocode,listings}
\usepackage{algorithmicx}
\usepackage{algpseudocode}
\usepackage{booktabs,multirow,adjustbox,diagbox,threeparttable,tabularray,setspace,wrapfig}
\usepackage{caption}
\usepackage{listings,colortbl,pifont}
\usepackage{placeins}
\usepackage{tabularray}
\UseTblrLibrary{booktabs}

\definecolor{codegreen}{rgb}{0,0.6,0}
\definecolor{codegray}{rgb}{0.5,0.5,0.5}
\definecolor{codepurple}{rgb}{0.58,0,0.82}
\definecolor{backcolour}{rgb}{1.0,1.0,1.0}
\lstdefinestyle{mystyle}{
    backgroundcolor=\color{backcolour},
    commentstyle=\color{codegreen},
    keywordstyle=\color{magenta},
    numberstyle=\tiny\color{codegray},
    stringstyle=\color{codepurple},
    basicstyle=\ttfamily\scriptsize,
    breakatwhitespace=false,
    breaklines=true,
    captionpos=b,
    keepspaces=true,
    numbers=left,
    numbersep=5pt,
    showspaces=false,
    showstringspaces=false,
    showtabs=false,
    tabsize=2,
    frame=single, % Adds a box around the code
    rulecolor=\color{black}
}
\definecolor{myred}{RGB}{213,123,101}
\definecolor{myblue}{RGB}{128,127,222}
\definecolor{mygreen}{RGB}{112,173,71}
\definecolor{cvprblue}{rgb}{0.21,0.49,0.74}
\usepackage[capitalize]{cleveref}  % Should be loaded after 'hyperref', and works perfectly with 'subfigure'.
\crefname{section}{Sec.}{Secs.}
\Crefname{section}{Section}{Sections}
\crefname{appendix}{App.}{Apps.}
\Crefname{appendix}{Appendix}{Appendices}
\crefname{table}{Tab.}{Tabs.}
\Crefname{table}{Table}{Tables}
\crefname{figure}{Fig.}{Figs.}
\Crefname{figure}{Figure}{Figures}
\crefname{equation}{Eq.}{Eqs.}
\Crefname{equation}{Equation}{Equations}
\crefname{theorem}{Thm.}{Thms.}
\Crefname{theorem}{Theorem}{Theorems}
\crefname{lemma}{Lem.}{Lems.}
\Crefname{lemma}{Lemma}{Lemmas}
\crefname{remark}{Rem.}{Rems.}
\Crefname{remark}{Remark}{Remarks}
\crefname{corollary}{Cor.}{Cors.}
\Crefname{corollary}{Corollary}{Corollaries}
\crefname{algorithm}{Alg.}{Algs.}
\Crefname{algorithm}{Algorithm}{Algorithms}
\newtheorem{theorem}{Theorem}
\newtheorem{corollary}{Corollary}[theorem]
\newtheorem{proposition}[theorem]{Proposition}
\newtheorem{appproposition}{Proposition}[section]
\theoremstyle{remark}
\newtheorem*{remark}{Remark}
\theoremstyle{plain}
\newcommand{\norm}[1]{\left\|#1\right\|}
\newcommand{\ip}[2]{\left\langle #1,#2\right\rangle}

\DeclareMathOperator{\Fix}{Fix}
\DeclareMathOperator{\dist}{dist}
\allowdisplaybreaks[1]

\title{Rectified Looped Transformer}

\author{%
Musheng Li \quad Jiawei Zhang \quad Ziyuan Liu \quad Yuantao Gu\thanks{Corresponding Author.} \\
Tsinghua University \\
\texttt{lms21@mails.tsinghua.edu.cn}
}

\newcommand{\tocite}[1]{{\color{red} [TO CITE]}}

\iclrfinalcopy % Uncomment for camera-ready version, but NOT for submission.
\begin{document}

\setcounter{footnote}{1}
\maketitle

\begin{abstract}
Looped Transformers increase effective depth by repeatedly applying a shared block, offering a parameter-efficient approach to language modeling.
However, most existing works directly reuse each loop's output as the next's input, which can lead to hidden-state scale growth and perturbation amplification as recurrence deepens. 
We propose Rectified Looped Transformer (RLT), which redesigns feedback between loops using an interpolation schedule inspired by flow-based modeling.
At each loop, RLT forms the next input by linearly interpolating between the current state and its prediction.
Such interpolation distributes state changes across loops, encouraging a smoother trajectory from the initial input to the final output.
Theoretical analysis proves that, under appropriate assumptions, RLT keeps state magnitudes uniformly bounded across loops and reduces the worst-case perturbation amplification.
We evaluate RLT on language modeling corpora and downstream benchmarks.
At matched training and inference depths, RLT consistently achieves lower perplexity than a standard looped Transformer, while exhibiting greater robustness to input perturbations and variance in inference depth.
Consistent with the theoretical results, latent trajectory analysis shows that RLT maintains more stable hidden-state norms across loops, which also facilitates extrapolation-based decoding strategies.
Notably, RLT achieves these gains without requiring additional trainable parameters or extra recurrent block evaluations, thus providing a simple yet more stable feedback mechanism for looped Transformers.

% Looped Transformers increase effective depth without adding physical parameters, offering a promising route to parameter-efficient language modeling. However, most existing approaches follow the Standard Looped Transformer (SLT) formulation, directly reusing each loop's output as the next input, which can lead to hidden-state scale growth and perturbation amplification as recurrence deepens. We propose Rectified Looped Transformer (RLT) with a feedback rule inspired by rectified flow. RLT treats each loop's output as an endpoint prediction and linearly interpolates it with the current input state to form the next input. By distributing the remaining displacement over the remaining loops, this rule separates endpoint prediction from state advancement. We theoretically prove that, under appropriate assumptions, RLT keeps state and output magnitudes uniformly bounded across loop depths and reduces the worst-case initial-perturbation amplification bound. Experiments show that RLT consistently achieves lower perplexity than SLT at matched training and inference depths, while improving robustness to numerical perturbations and inference with both fewer and more loops than used in training. Latent-trajectory analyses reveal more controlled state evolution, with improved scale stability also benefiting extrapolation-based decoding. Notably, these gains require only elementwise interpolation, with no additional trainable feedback parameters or recurrent-block forward passes.

\end{abstract}

\section{Introduction}
\label{sec:introduction}

Improving language model capabilities while controlling parameter count is an important challenge in model scaling. Looped Transformers address this challenge by repeatedly applying a shared stack of layers, increasing effective computational depth without increasing the number of physical parameters. This makes recurrence depth an additional axis for scaling language models. Early work~\citep{saunshi2025reasoning} establishes the reasoning potential of looped Transformers through controlled experiments and theoretical analysis. Huginn~\citep{geiping2025scaling} demonstrates their feasibility at scale, while Ouro~\citep{zhu2025scaling} further scales pretraining to trillions of tokens, achieving performance competitive with larger standard LLMs and gains on tasks requiring multi-hop reasoning. Together, these advances establish latent recurrent computation as a promising route to more capable, parameter-efficient language models.

A key design choice in recurrent computation is how each loop's output becomes the next loop's input. GRUs~\citep{cho2014learning} use interpolation between existing and candidate states for effective sequence modeling~\citep{chung2014empirical}; \emph{Thinking with Looped Flows}~\citep{suleymanzade2026thinking} couples answer-space flow updates with recurrence to improve structured reasoning. In the direct-feedback formulation (hereafter referred to as Standard Looped Transformer, SLT) commonly used in current looped language models, including Ouro~\citep{zhu2025scaling}, Loopie~\citep{gao2026loop}, and LoopFormer~\citep{jeddi2026loopformer}, the network output fully replaces the current input, without retaining it as a separate component of the feedback update. From a dynamical systems perspective, this applies the entire change from the current state to the network prediction at every loop, coupling prediction with state advancement. As the shared network is repeatedly applied, these changes can accumulate into hidden-state scale growth and perturbation amplification. STARS~\citep{yang2026stabilizing} links performance degradation at larger recurrent depths to unstable latent dynamics; our SLT likewise exhibits sensitivity to loop count, accompanied by pronounced changes in state scale. This raises a concrete question: can retaining the current input state and controlling its advance toward each new prediction yield more effective and stable recurrent computation?

\begin{figure}[t]
    \centering
    \includegraphics[width=0.98\linewidth]{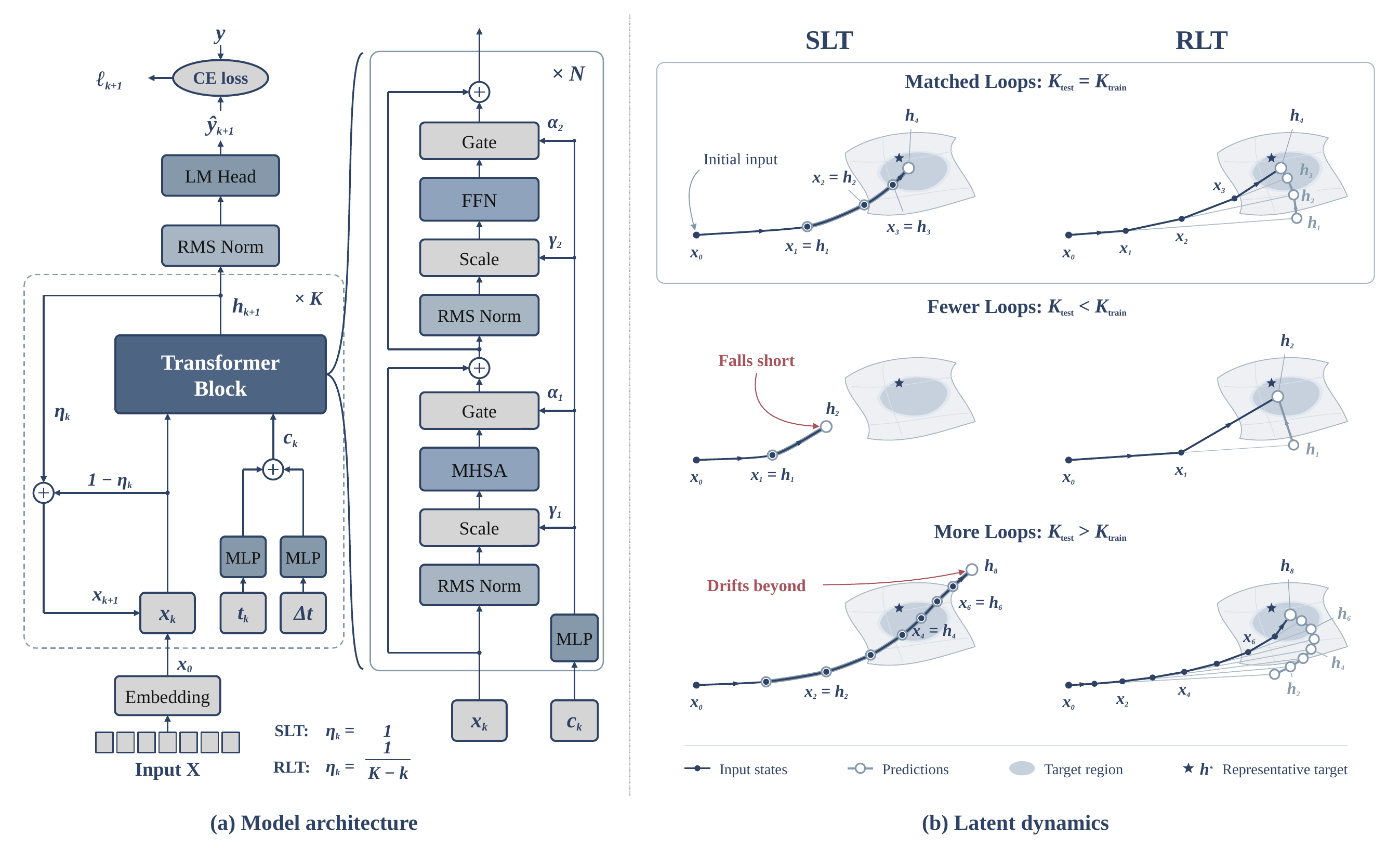}
    \caption{\textbf{Overview of Rectified Looped Transformer (RLT).} \textbf{(a)} SLT and RLT share the same looped architecture and uniform all-step supervision. RLT replaces direct feedback with interpolation between the current input and the predicted endpoint.
    \textbf{(b)} Schematic latent dynamics under matched and mismatched loop counts, illustrating more uniform state advancement and more stable final predictions in RLT, while SLT may fall short or drift beyond the target region.}
    \label{fig:rlt-overview}
    \vspace{-0.5em}
\end{figure}

To address this gap, we propose Rectified Looped Transformer (RLT; Figure~\ref{fig:rlt-overview}(a)), which organizes recurrent feedback through linear interpolation inspired by rectified flow~\citep{liu2023flow}. RLT treats the network output at each loop as an endpoint prediction and combines it with the current input state to form the next input. The interpolation weight on the endpoint prediction increases as the remaining number of loops decreases. Intuitively, each loop re-estimates the endpoint and distributes the remaining displacement over the remaining steps. This provides a geometric intuition of approximately linear progress toward a shared target region, while allowing the endpoint estimate to evolve across loops (Figure~\ref{fig:rlt-overview}(b)). The design separates prediction from state advancement and explicitly organizes how network updates accumulate across loops. Our theory establishes depth-uniform bounds on feedback-state and output magnitudes under time-dependent residual control. Even when the network can amplify perturbations, RLT reduces the worst-case amplification bound for initial perturbations from exponential to polynomial in depth under a uniform sensitivity bound.

Empirically, with identical uniform all-step supervision, RLT consistently achieves lower perplexity than the direct-feedback SLT when training and inference depths are matched. Besides, experiments show greater robustness to perturbations and changes in inference depth, including inference with both fewer and more loops than used during training. Trajectory analyses reveal differences in scale accumulation and prediction evolution, while improved scale stability also benefits contrastive decoding and guidance. Notably, these gains require only elementwise interpolation, with no additional trainable feedback parameters or recurrent-block forward passes, highlighting the efficiency of RLT’s feedback design.

Our main contributions are as follows:
\begin{itemize}
    \item We introduce Rectified Looped Transformer (RLT), which interprets each loop's output as an endpoint prediction and organizes latent-state feedback through a remaining-time interpolation schedule.
    \item We theoretically analyze state and prediction magnitudes and perturbation amplification, establishing depth-uniform magnitude bounds under time-dependent residual control and polynomial bounds on initial-perturbation amplification under a uniform Lipschitz condition.
    \item We empirically demonstrate improvements in language modeling and robustness to changes in loop count and numerical perturbations, supported by latent-trajectory analyses and extrapolation-based decoding experiments.
\end{itemize}
\section{Related Work}
\label{sec:related_work}

\paragraph{Looped Transformers for Language Modeling.}
Looped Transformers repeatedly apply a shared stack of layers, increasing effective computational depth without adding physical parameters. Early architectures such as Universal Transformers~\citep{dehghani2019universal} and ALBERT~\citep{lan2020albert} established parameter sharing across depth, while subsequent theoretical and empirical studies demonstrated its potential for algorithm execution~\citep{giannou2023} and reasoning~\citep{saunshi2025reasoning}. Recently, Huginn~\citep{geiping2025scaling}, Ouro~\citep{zhu2025scaling}, and Loopie~\citep{gao2026loop} have extended this paradigm to large-scale language modeling, achieving performance competitive with larger standard LLMs on commonsense and reasoning benchmarks while using fewer physical parameters. Existing efforts seek to improve the performance and efficiency of looped language models through sparse experts~\citep{gao2026loop,chen2026loopmoe,lee2026sparse}, depth-specific adapters~\citep{bae2025relaxed}, and adaptive allocation of recurrent computation~\citep{bae2025mixture,fu2026think,you2026improving}, etc. In this work, we focus on state feedback and propose RLT to improve language modeling and recurrent stability.

\paragraph{State Feedback and Recurrent Stability.}
State feedback determines how each loop's output becomes the next input. The most common practice in looped Transformers is direct feedback, which reuses the output as the next recurrent state, as in Ouro~\citep{zhu2025scaling}, Loopie~\citep{gao2026loop}, and LoopFormer~\citep{jeddi2026loopformer}. Recent work has begun to explore feedback control by scaling the residual update between the current state and the network output. ELR~\citep{ladhak2026one} seeks to preserve one-loop performance while benefiting from additional loops, using learned gates to scale residual updates; TAPS~\citep{wang2026taps} aims to improve recurrent inference efficiency by adaptively adjusting update magnitudes according to persistent progress and fluctuation estimated from the trajectory. Recent studies such as STARS~\citep{yang2026stabilizing} also reveal that looped Transformers can become unstable as recurrence deepens. To stabilize these dynamics, Fixed-Point Reasoners~\citep{movahedi2026} use damped feedback to suppress oscillation and promote convergence toward fixed points. Our RLT uses a predefined remaining-time interpolation schedule, with theoretical analysis establishing tighter control over state magnitudes and perturbation amplification than direct feedback.

\paragraph{Flow-based Modeling and Iterative Reasoning.}
Flow-based generative modeling~\citep{lipman2023} connects individual state updates to continuous transport between distributions through learned ODEs. Rectified Flow~\citep{liu2023flow} uses linear interpolation between source and target samples as training paths, encouraging straighter flows that support sampling with fewer steps. Recent work combines looped computation with flow-based generation. In structured reasoning tasks such as Sudoku and maze solving, \textit{Thinking with Looped Flows}~\citep{suleymanzade2026thinking} couples a noise-to-answer flow with recurrent hidden states and local denoising supervision; \textit{Unrolled Flow Models}~\citep{izermine2026} use a DiT~\citep{peebles2023scalable} backbone to learn flow-based answer generation through their own latent rollouts with a terminal prediction loss. In contrast, our RLT applies flow-inspired interpolation to state feedback in autoregressive language models, combining each loop's output with its input state to form the next input.
\section{Rectified Looped Transformer}
\label{sec:rlt}

\subsection{Preliminaries: Looped Computation and State Feedback}
\label{subsec:looped_computation_and_state_feedback}

As illustrated in Figure~\ref{fig:rlt-overview}(a), a looped Transformer repeatedly applies a shared stack of layers to the input sequence's representation. We focus on the recurrent computation for predicting the next token from a fixed input context. The initial input state at the prediction position is $\mathbf x_0\in\mathbb R^d$, formed from token and positional embeddings, where $d$ is the hidden size. We associate $K\ge1$ loops with a uniform time grid on $[0,1]$:
\begin{align}
    \Delta t=\frac1K,\qquad t_k=\frac{k}{K},\qquad k=0,\ldots,K-1.
    \label{eq:uniform_time_grid}
\end{align}
The shared Transformer stack $\mathbf F_\theta$ performs one loop of computation and receives time and step-size conditioning through AdaLN-style modulation~\citep{peebles2023scalable}. For brevity, we suppress its step-size argument and conditioning on the current prefix representations below. At loop $k$, $\mathbf x_k$ denotes the input state and $\mathbf h_{k+1}$ the hidden output. A feedback coefficient $\eta_k$ determines their contribution to the next input:
\begin{align}
    \mathbf h_{k+1}=\mathbf F_\theta(\mathbf x_k,t_k),\qquad \mathbf x_{k+1}=(1-\eta_k)\mathbf x_k+\eta_k\mathbf h_{k+1},\qquad 0\le\eta_k\le1.
    \label{eq:update}
\end{align}
Standard Looped Transformer (SLT) uses direct feedback, setting $\eta_k=1$ and directly feeding the output $\mathbf h_{k+1}$ back as the next input state. Parameters are shared across loops and held fixed when varying inference depth.

The shared readout $\mathcal D$ acts on each hidden output:
\begin{align}
    \mathcal D(\mathbf h)=\mathbf W\operatorname{RMSNorm}(\mathbf h),\qquad \ell_k=\operatorname{CE}(\mathcal D(\mathbf h_k),\mathbf y),
    \label{eq:readout_supervision}
\end{align}
where $\mathbf W$ is the vocabulary projection, $\mathbf y$ denotes the next-token target, and $\operatorname{CE}$ is the cross-entropy loss. In this work, the looped Transformer receives uniform all-step supervision with training loss $K^{-1}\sum_{k=1}^K\ell_k$. Inference decodes the final hidden output $\mathbf h_K$.

\subsection{Flow-inspired State Feedback}
\label{subsec:flow_inspired_state_feedback}

Inspired by linear interpolation in rectified flow~\citep{liu2023flow}, we propose Rectified Looped Transformer (RLT), which treats each loop's network output as an endpoint prediction and separates prediction from state advancement. At time $t_k$, the displacement from the current state to the predicted endpoint is $\mathbf h_{k+1}-\mathbf x_k$, and the remaining time is $1-t_k$. Advancing by one time step gives
\begin{equation*}
    \mathbf x_{k+1}=\mathbf x_k+\Delta t\,\frac{\mathbf h_{k+1}-\mathbf x_k}{1-t_k}.
\end{equation*}
The feedback weight is therefore the fraction of the remaining time covered by one step:
\begin{align}
    \eta_k=\frac{\Delta t}{1-t_k}=\frac1{K-k},\qquad k=0,\ldots,K-1.
    \label{eq:schedule}
\end{align}
The weight increases from $\eta_0=1/K$ to $\eta_{K-1}=1$, forming a smoothly interpolated trajectory from $\mathbf x_0$ to the final prediction $\mathbf x_K=\mathbf h_K$. Intuitively, each loop re-estimates the endpoint and distributes the remaining displacement over the remaining loops. When endpoint estimates agree, the rule yields uniformly spaced states along the line from the initial state to the endpoint (Figure~\ref{fig:rlt-overview}(b)).

We next establish stability guarantees for RLT by bounding state and output norms and initial-perturbation amplification as loop depth increases.

\subsection{Theoretical Analysis of Stability}
\label{subsec:theoretical_analysis_of_stability}

We establish depth-uniform bounds on state and hidden-output norms and polynomial bounds on initial-perturbation amplification for the RLT schedule. Throughout, $\|\cdot\|$ and $\langle\cdot,\cdot\rangle$ denote the Euclidean norm and inner product. Constants in depth-uniform bounds, including those hidden by $\lesssim$, are independent of $K$. Theorem~\ref{thm:bounded} below bounds state norms under general feedback schedules.
\begin{theorem}
    \label{thm:bounded}
    Suppose there exist $M\ge0$ and $0<\alpha\le1$ such that, for all $0\le t<1$,
    \begin{align}
        \sup_{\mathbf x\in\mathbb R^d}\|\mathbf F_\theta(\mathbf x,t)-\mathbf x\|\le M(1-t)^\alpha.
        \label{eq:decay}
    \end{align}
    Then the feedback rule in Eq.~(\ref{eq:update}) satisfies
    \begin{align}
        \|\mathbf x_k\|\le\|\mathbf x_0\|+M\sum_{j=0}^{k-1}\eta_j(1-t_j)^\alpha,\qquad 1\le k\le K.
        \label{eq:generalbound}
    \end{align}
\end{theorem}

\begin{remark}
    For $\alpha=1$, Eq.~(\ref{eq:decay}) is equivalent to $\sup_{\mathbf x,t}\|\mathbf v_\theta(\mathbf x,t)\|\le M$ for the flow-inspired velocity field
    \begin{align}
        \mathbf v_\theta(\mathbf x,t)=\frac{\mathbf F_\theta(\mathbf x,t)-\mathbf x}{1-t}.
        \label{eq:flow_velocity}
    \end{align}
    For $0<\alpha<1$, the condition allows $\sup_{\mathbf x}\|\mathbf v_\theta(\mathbf x,t)\|\le M(1-t)^{\alpha-1}$, relaxing uniform velocity boundedness to permit integrable growth near $t=1$.
\end{remark}

In practice, we observe pronounced norm growth under SLT's direct feedback, while RLT maintains more stable norms (Section~\ref{subsec:recurrent_stability}). The following two corollaries characterize the different depth dependence of these feedback rules.
\begin{corollary}
    \label{cor:directstate}
    Under the assumptions of Theorem~\ref{thm:bounded}, SLT's direct feedback ($\eta_k\equiv1$) gives
    \begin{equation}
        \|\mathbf x_K\|\le\|\mathbf x_0\|+M\underbrace{K^{-\alpha}\sum_{j=1}^Kj^\alpha}_{\longrightarrow\,\infty\quad(K\to\infty)}.
        \label{eq:directstate}
    \end{equation}
\end{corollary}

\begin{corollary}
    \label{cor:bounded}
    Under the assumptions of Theorem~\ref{thm:bounded}, the RLT schedule in Eq.~(\ref{eq:schedule}) gives
    \begin{equation}
        \max\left\{\max_{0\le k\le K}\|\mathbf x_k\|,\max_{1\le k\le K}\|\mathbf h_k\|\right\}\le\underbrace{\|\mathbf x_0\|+\frac M\alpha}_{\text{constant in }K}.
        \label{eq:bothbound}
    \end{equation}
\end{corollary}

These corollaries show that SLT's direct feedback permits unbounded state-norm growth with depth (Appendix~\ref{app:proof_of_thm1}), whereas the RLT schedule keeps both input-state and hidden-output norms uniformly bounded as loop depth increases. Under an additional radial condition on the residual, we obtain the following stronger bound.
\begin{corollary}
    \label{cor:radial}
    Under the assumptions of Theorem~\ref{thm:bounded} and the RLT schedule, suppose $\langle\mathbf x_k,\mathbf h_{k+1}-\mathbf x_k\rangle\le0$ for $0\le k<K$. Then
    \begin{align}
        \max_{0\le k\le K}\|\mathbf x_k\|^2\le\|\mathbf x_0\|^2+M^2\underbrace{K^{-2\alpha}\sum_{j=1}^Kj^{2\alpha-2}}_{\longrightarrow\,0\quad(K\to\infty)}.
        \label{eq:radialbound}
    \end{align}
\end{corollary}
The radial condition makes the first-order contribution to squared norm growth nonpositive, while the accumulated squared update magnitudes vanish with depth. Proofs of Theorem~\ref{thm:bounded} and its corollaries are provided in Appendix~\ref{app:proof_of_thm1}.

We next examine how perturbations to the initial state $\mathbf x_0$ propagate across loops. The following theorem bounds the separation between two trajectories under a Lipschitz condition on the shared network.
\begin{theorem}
    \label{thm:perturbation}
    Suppose $\mathbf F_\theta(\mathbf x,t_k)$ is $L_k$-Lipschitz in $\mathbf x$, with $L_k\ge0$. If $\mathbf x_k$ and $\widetilde{\mathbf x}_k$ follow Eq.~(\ref{eq:update}) with the same contextual conditioning and feedback schedule, then
    \begin{align}
        \|\mathbf x_k-\widetilde{\mathbf x}_k\|\le\prod_{j=0}^{k-1}(1-\eta_j+\eta_jL_j)\|\mathbf x_0-\widetilde{\mathbf x}_0\|,\qquad 1\le k\le K.
        \label{eq:product}
    \end{align}
\end{theorem}

Theorem~\ref{thm:perturbation} bounds trajectory separation through a product of per-loop factors, relating the difference between terminal states to the initial perturbation. A uniform Lipschitz bound $L>1$ allows perturbations to be amplified across loops. The following corollary compares the resulting depth dependence under the feedback rules of RLT and SLT.
\begin{corollary}
    \label{cor:amplification}
    Under the assumptions of Theorem~\ref{thm:perturbation}, suppose $L_k\le L$ uniformly in $k$ and $K$, with $L>1$. Then SLT's direct feedback ($\eta_k\equiv1$) gives
    \begin{align}
        \|\mathbf x_K-\widetilde{\mathbf x}_K\|\lesssim L^K\|\mathbf x_0-\widetilde{\mathbf x}_0\|.
        \label{eq:directperturb}
    \end{align}
    For the RLT schedule,
    \begin{align}
        \|\mathbf x_K-\widetilde{\mathbf x}_K\|\lesssim K^{L-1}\|\mathbf x_0-\widetilde{\mathbf x}_0\|.
        \label{eq:polynomial}
    \end{align}
\end{corollary}
Corollary~\ref{cor:amplification} shows that the RLT schedule reduces the depth dependence of the worst-case amplification bound from exponential to polynomial, providing tighter control over sensitivity to initial perturbations. Since $\mathbf x_K=\mathbf h_K$, it also controls the sensitivity of RLT's final hidden output. Proofs of Theorem~\ref{thm:perturbation} and Corollary~\ref{cor:amplification} are provided in Appendix~\ref{app:proof_of_thm2}.

For a time-independent nonexpansive network, recurrent computation can also be viewed as fixed-point iteration~\citep{movahedi2026}. RLT then follows the classical Krasnosel'skii--Mann iteration~\citep{cominetti2014}, yielding the following fixed-point residual bound.
\begin{proposition}
    \label{prop:fixedpoint}
    Suppose $\mathbf F_\theta(\mathbf x,t)=\mathbf F_\theta(\mathbf x)$ is nonexpansive on $\mathbb R^d$ and $\operatorname{Fix}(\mathbf F_\theta):=\{\mathbf x\in\mathbb R^d:\mathbf F_\theta(\mathbf x)=\mathbf x\}\ne\varnothing$. Then, for the RLT schedule and $K>2$,
    \begin{align}
        \|\mathbf F_\theta(\mathbf x_K)-\mathbf x_K\|\lesssim\frac{\operatorname{dist}(\mathbf x_0,\operatorname{Fix}(\mathbf F_\theta))}{\sqrt{\log K}},
        \label{eq:fixedresidual}
    \end{align}
    where $\operatorname{dist}$ denotes distance from a point to a set.
\end{proposition}
Proposition~\ref{prop:fixedpoint} follows from \citet[Proposition~11]{cominetti2014}; a brief proof appears in Appendix~\ref{app:proof_of_prop3}. Together, these results connect RLT's remaining-time feedback rule to controlled state scales, reduced perturbation amplification bounds, and fixed-point residual decay. These properties provide theoretical support for the stability of RLT's recurrent computation.
% Full-width figures occupy the top area; float-only pages also align at top.
\makeatletter
\setlength{\@fptop}{0pt}
\setlength{\@fpsep}{12pt plus 2pt minus 2pt}
\makeatother
\setlength{\floatsep}{8pt plus 2pt minus 2pt}
\setlength{\textfloatsep}{10pt plus 2pt minus 2pt}

\section{Experiments}
\label{sec:experiments}

We compare SLT and RLT in language modeling, recurrent stability, and latent-state dynamics. Matched architectures and training settings allow us to examine the effect of state feedback.

Following LoopFormer~\citep{jeddi2026loopformer}, we adopt a GPT-2-style looped Transformer conditioned on time and step size ({Figure~\ref{fig:rlt-overview}}(a)). Each loop maps its input state to a hidden output through the shared Transformer stack. SLT feeds this hidden output back as the next loop's input state, whereas RLT constructs the next input by interpolating between the current input state and the hidden output using a remaining-time schedule. These feedback rules are formalized in Sections~\ref{subsec:looped_computation_and_state_feedback}--\ref{subsec:flow_inspired_state_feedback}. Our default configuration is $3\otimes8$: three shared Transformer layers applied over eight loops, with 278M physical parameters. We denote the training and inference loop counts by $K_{\mathrm{train}}$ and $K_{\mathrm{test}}$, respectively. Unless otherwise specified, inference follows the uniform time grid defined in Eq.~(\ref{eq:uniform_time_grid}), with $t_k=k/K_{\mathrm{test}}$ and $\Delta t=1/K_{\mathrm{test}}$.

Both models are trained from scratch on the deduplicated version of The Pile~\citep{gao2020pile}, with the training data comprising 6.88B tokens and uniform all-step next-token prediction supervision. We train for 14,000 optimizer steps with a peak learning rate of $6\times10^{-4}$ and a global batch size of 480 sequences of 1,024 tokens. We measure perplexity on The Pile~\citep{gao2020pile}, FineWeb-Edu (FWE)~\citep{penedo2024fineweb}, and OpenWebText (OWT)~\citep{gokaslan2019openwebtext}, and evaluate zero-shot accuracy following \texttt{lm-eval harness}~\citep{gao2023framework}. Our downstream evaluation covers ten well-established benchmarks spanning commonsense reasoning, reading comprehension, scientific question answering, and discourse completion: COPA~\citep{roemmele2011choice}, HellaSwag (HS)~\citep{zellers2019hellaswag}, LAMBADA (LB)~\citep{paperno2016lambada}, OpenBookQA (OBQA)~\citep{mihaylov2018suit}, PIQA~\citep{bisk2020piqa}, RACE~\citep{lai2017race}, SciQ~\citep{welbl2017crowdsourcing}, ARC~\citep{clark2018think}, Social IQA (SIQA)~\citep{sap2019social}, and WinoGrande (WG)~\citep{sakaguchi2021winogrande}. We report accuracy or normalized accuracy as appropriate, averaging ARC-Easy and ARC-Challenge before computing the ten-task mean.

Further details on the model architecture, training hyperparameters, and datasets are provided in Appendices~\ref{app:model_architecture}, \ref{app:training_implementation}, and~\ref{app:data_evaluation}, respectively.

% ----------------------------------------------------------- %

\subsection{Language Modeling Performance}
\label{subsec:language_modeling_performance}

We first compare SLT and RLT at matched training and inference depths, $K_{\mathrm{train}}=K_{\mathrm{test}}=K\in\{2,4,8\}$. At each depth, both models use three shared Transformer layers, identical training settings, and uniform all-step supervision. Table~\ref{tab:matched-depth} shows that RLT consistently achieves lower perplexity across all three depths and all three evaluation corpora. At $K=4$, it reduces PPL by 4.2\%, 5.5\%, and 5.2\% on Pile, FWE, and OWT, respectively. This advantage also holds at $K=2$ and $K=8$.

% Uses the existing table packages: booktabs, multirow, xcolor, colortbl, and fix-cm.
\begin{table}[t]
\centering
\caption{Language modeling and downstream performance at matched depth $K=K_{\mathrm{train}}=K_{\mathrm{test}}$. Structure denotes shared layers $\otimes$ loops. Accuracy is reported in percentages; Avg Acc is the unweighted mean across ten benchmarks. \textbf{Bold} marks the better result within each depth group.}
\label{tab:matched-depth}
\begingroup
\definecolor{rltgroup}{HTML}{F0F1F3}
\definecolor{rltrow}{HTML}{EBF3FA}
\definecolor{rltdivider}{HTML}{9298A0}
\setlength{\tabcolsep}{1.75pt}
\setlength{\aboverulesep}{1.2pt}
\setlength{\belowrulesep}{1.2pt}
\renewcommand{\arraystretch}{1.22}
\fontsize{8}{9.6}\selectfont
\begin{tabular}{@{}c!{\color{rltdivider}\vrule width .35pt}c!{\color{rltdivider}\vrule width .35pt}ccc!{\color{rltdivider}\vrule width .35pt}cccccccccc!{\color{rltdivider}\vrule width .35pt}c@{}}
\toprule
\multirow{2}{*}{\textbf{Variants}} & \multirow{2}{*}{\textbf{Structure}} & \multicolumn{3}{c!{\color{rltdivider}\vrule width .35pt}}{\textbf{Perplexity} $\downarrow$} & \multicolumn{11}{c}{\textbf{Accuracy (\%)} $\uparrow$} \\
\cmidrule(lr){3-5}\cmidrule(l){6-16}
 & & \textbf{Pile} & \textbf{FWE} & \textbf{OWT} & \textbf{COPA} & \textbf{HS} & \textbf{LB} & \textbf{OBQA} & \textbf{PIQA} & \textbf{RACE} & \textbf{SciQ} & \textbf{ARC} & \textbf{SIQA} & \textbf{WG} & \textbf{Avg Acc} \\
\midrule
\rowcolor{rltgroup}
\multicolumn{16}{c}{$\boldsymbol{K = 8}$} \\
\midrule
SLT & $3 \otimes 8$ & 11.08 & 27.58 & 25.39 & \textbf{64.00} & \textbf{36.50} & 33.00 & \textbf{29.50} & \textbf{62.50} & 28.57 & 66.50 & 30.22 & 36.92 & \textbf{55.91} & \textbf{44.36} \\
\rowcolor{rltrow}
RLT & $3 \otimes 8$ & \textbf{10.82} & \textbf{26.74} & \textbf{24.61} & 62.00 & \textbf{36.50} & \textbf{34.00} & 28.00 & 60.87 & \textbf{29.52} & \textbf{70.50} & \textbf{31.31} & \textbf{38.46} & 44.09 & 43.53 \\
\midrule
\rowcolor{rltgroup}
\multicolumn{16}{c}{$\boldsymbol{K = 4}$} \\
\midrule
SLT & $3 \otimes 4$ & 11.76 & 29.76 & 27.30 & 64.00 & \textbf{35.50} & 26.00 & 27.50 & 59.78 & \textbf{26.67} & \textbf{66.50} & 32.38 & \textbf{37.44} & 48.03 & 42.38 \\
\rowcolor{rltrow}
RLT & $3 \otimes 4$ & \textbf{11.27} & \textbf{28.14} & \textbf{25.88} & \textbf{68.00} & \textbf{35.50} & \textbf{31.00} & \textbf{29.00} & \textbf{60.33} & 23.81 & 65.50 & \textbf{33.01} & 34.87 & \textbf{51.97} & \textbf{43.30} \\
\midrule
\rowcolor{rltgroup}
\multicolumn{16}{c}{$\boldsymbol{K = 2}$} \\
\midrule
SLT & $3 \otimes 2$ & 12.49 & 32.24 & 29.52 & \textbf{64.00} & \textbf{35.50} & 23.50 & \textbf{27.50} & 59.24 & 20.95 & \textbf{65.50} & \textbf{31.94} & \textbf{39.49} & \textbf{48.03} & \textbf{41.56} \\
\rowcolor{rltrow}
RLT & $3 \otimes 2$ & \textbf{12.44} & \textbf{32.04} & \textbf{29.33} & 60.00 & 35.00 & \textbf{28.50} & 25.50 & \textbf{60.87} & \textbf{22.86} & 63.00 & 29.81 & \textbf{39.49} & \textbf{48.03} & 41.31 \\
\bottomrule
\end{tabular}
\endgroup
\end{table}

\begin{figure}[t]
\centering
\includegraphics[width=\linewidth]{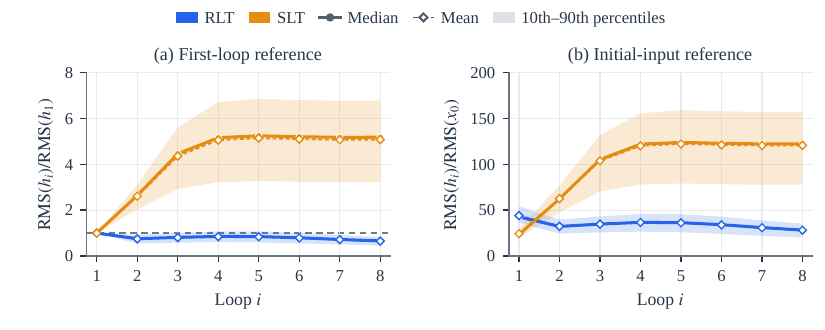}
\caption{Cross-loop hidden-state scales relative to the first-loop output $\mathbf{h}_1$ (left) and the initial embedding input $\mathbf{x}_0$ (right), with $K_{\mathrm{train}}=K_{\mathrm{test}}=8$.}
\label{fig:stability_scale}
\end{figure}

On downstream benchmarks, the two methods achieve comparable average accuracy, with task-specific differences across depths. At $K=4$, RLT increases the ten-task average from 42.38\% to 43.30\%. LAMBADA accuracy improves at all three depths, with gains of 5 percentage points at both $K=2$ and $K=4$. These results show that reorganizing state feedback consistently improves language modeling while maintaining competitive downstream performance. We next examine how this feedback design affects the numerical behavior of recurrent computation.

% ----------------------------------------------------------- %

\subsection{Recurrent Stability and Robustness}
\label{subsec:recurrent_stability}

We first examine hidden-state scales and responses to initial-input and recurrent perturbations to study RLT's recurrent stability. We then assess the practical effects of these properties through contrastive decoding and guidance, followed by an evaluation of performance under quantization.

\paragraph{Hidden-state scale.} We measure the scale of each loop's output $\mathbf{h}_i$ before the final RMS normalization. Figure~\ref{fig:stability_scale} compares the token-wise RMS ratios relative to the first-loop output $\mathbf{h}_1$ and the initial embedding input $\mathbf{x}_0$. SLT exhibits pronounced scale accumulation over the early loops: by loop eight, the mean ratio relative to $\mathbf{h}_1$ reaches 5.08, compared with 0.65 for RLT. The comparison relative to $\mathbf{x}_0$ likewise shows substantial scale accumulation in SLT, while RLT maintains a more controlled scale across loops. These observations are consistent with the output-magnitude bound in Corollary~\ref{cor:bounded}. Measurement details are provided in Appendix~\ref{app:stability_scale}.

\begin{figure}[t]
\centering
\includegraphics[width=\linewidth]{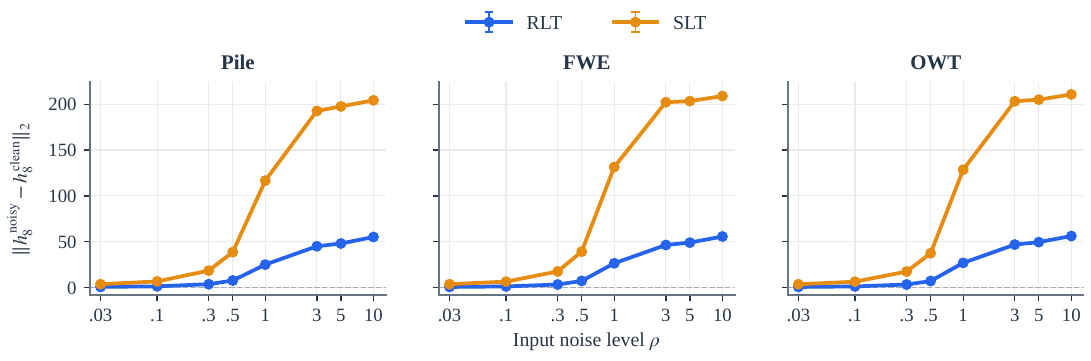}
\caption{Final-output displacement after injecting Gaussian noise into the initial embedding input $\mathbf{x}_0$. Distances are averaged over tokens and measured before RMS normalization.}
\label{fig:stability_initial_noise}
\end{figure}

\begin{figure}[t]
\centering
\includegraphics[width=\linewidth]{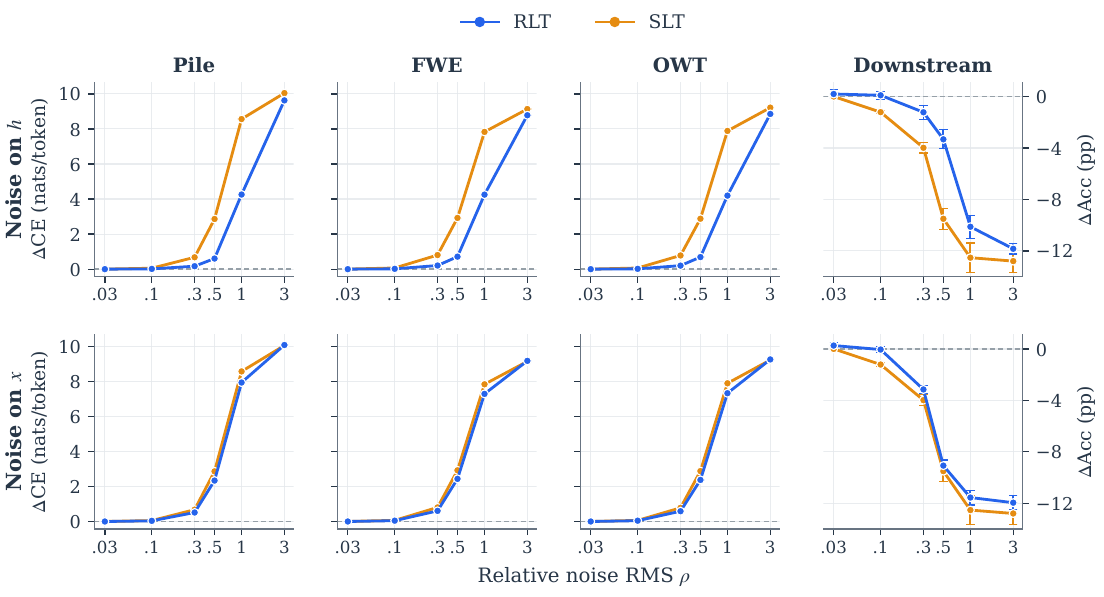}
\caption{Performance under recurrent perturbations applied to $\mathbf{h}$ before feedback (top) or $\mathbf{x}$ after feedback (bottom). $\Delta\mathrm{CE}$ and $\Delta\mathrm{Acc}$ are changes relative to each model's clean baseline.}
\label{fig:stability_recurrent_noise}
\end{figure}

\paragraph{Robustness to perturbations.} We first inject RMS-scaled Gaussian noise into the initial embedding input $\mathbf{x}_0$ and measure the resulting displacement of the final hidden output. Across all three corpora and all tested noise levels, RLT exhibits smaller absolute output displacements (Figure~\ref{fig:stability_initial_noise}), in line with the initial-perturbation analysis in Theorem~\ref{thm:perturbation}.

We then inject noise at every loop, either into the block output $\mathbf{h}$ before feedback or into the state $\mathbf{x}$ after feedback. As shown in Figure~\ref{fig:stability_recurrent_noise}, RLT better preserves predictive performance overall under both schemes, with larger benefits for output perturbations. At $\rho=0.5$, output perturbations increase Pile cross-entropy by 0.61 nats/token for RLT versus 2.86 for SLT. The dependence on injection location reflects the role of interpolation: output noise passes through the feedback coefficient, whereas noise added after feedback enters the next loop directly. See Appendix~\ref{app:stability_noise} for experimental details.

\begin{figure}[t]
\centering
\captionsetup{font=small,skip=5pt,justification=raggedright,singlelinecheck=false}
\begin{minipage}[t]{0.66\linewidth}
\vspace{0pt}
\centering
\includegraphics[width=\linewidth]{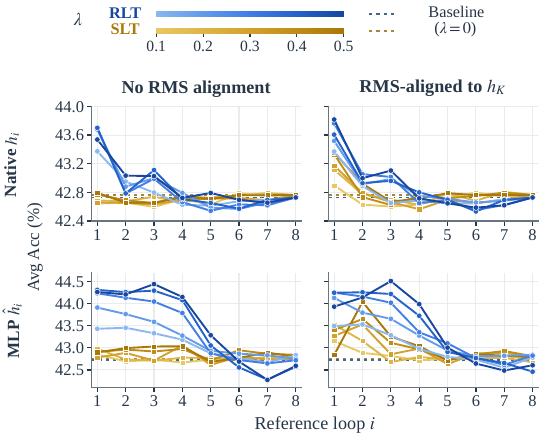}
\caption{Loop-wise contrastive decoding with native outputs $\mathbf{h}_i$ (top) or MLP predictions $\hat{\mathbf{h}}_i$ (bottom). References retain their original scale (left) or are RMS-aligned to $\mathbf{h}_K$ (right). Color intensity indicates extrapolation strength $\lambda$; horizontal dashed lines mark unguided baselines.}
\label{fig:stability_loopcd}
\end{minipage}\hfill
\begin{minipage}[t]{0.32\linewidth}
\vspace{0pt}
\centering
\includegraphics[width=\linewidth]{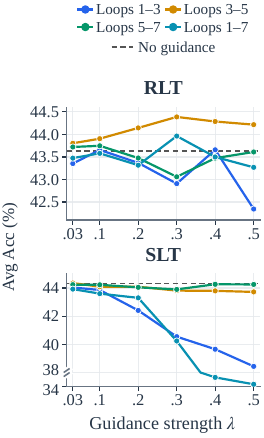}
\caption{Guidance applied over different loop intervals for RLT and SLT. Colors indicate the guided loops, and dashed lines mark unguided baselines.}
\label{fig:stability_guidance}
\end{minipage}
\end{figure}

\paragraph{Loop-wise contrastive decoding.} Contrastive decoding~\citep{li2023contrastive} exploits differences between stronger and weaker predictions to improve downstream task performance; related methods in looped Transformers~\citep{yu2026loopcd,liu2026decoding} construct these contrasts from different recurrent depths. We apply hidden-state extrapolation before the final readout, using either native intermediate outputs $\mathbf{h}_i$ or predictions $\hat{\mathbf{h}}_i$ from an auxiliary MLP branch. The branch predicts the complete block output from its first layer's representation and is trained with relative mean squared error (see Appendix~\ref{app:stability_loopcd} for details).

Using references from the first three loops, RLT obtains consistently larger accuracy gains than SLT across both reference types and RMS-alignment settings. As shown in Figure~\ref{fig:stability_loopcd}, RLT benefits more from early references without RMS alignment, whereas SLT shows little or no improvement. Aligning reference magnitudes to the final output makes SLT's early references more effective: with the first-loop reference and $\lambda=0.5$, accuracy increases from 42.79\% to 43.35\%. RLT also benefits from MLP references, reaching 44.45\% from a 42.73\% baseline with the third-loop reference at $\lambda=0.5$ without alignment. These findings highlight how RLT's more stable representation scales support the effective use of intermediate predictions and improve contrastive decoding.

% Measure both columns together so the surrounding text and narrow figure cannot split.
\FloatBarrier
\noindent\begin{minipage}{\linewidth}
\begin{minipage}[t]{\dimexpr0.40\linewidth-12pt\relax}
\vspace{0pt}
\paragraph{Guidance during recurrence.} We further examine guidance within the recurrent computation, drawing on autoguidance with a weaker version of the model~\citep{karras2024guiding}. At selected loops, we extrapolate the block output away from the RMS-aligned MLP prediction and feed the result into subsequent computation. See Appendix~\ref{app:stability_guidance} for details. Guiding loops 3--5 improves RLT at every tested strength shown in Figure~\ref{fig:stability_guidance}, reaching a gain of 0.76 percentage points at $\lambda=0.3$. In SLT, guidance over loops 1--3 or 1--7 causes substantial degradation as strength increases. RLT can therefore benefit from interventions during recurrence, with the middle loops providing an effective guidance interval.
\end{minipage}\hfill
\begin{minipage}[t]{0.60\linewidth}
\vspace{0pt}
\centering
\includegraphics[width=\linewidth]{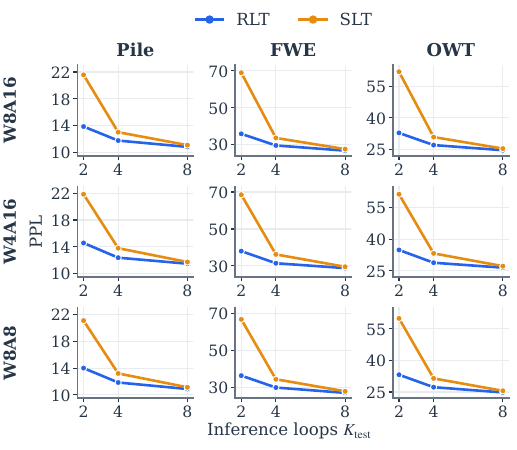}
\captionsetup{font=small,skip=4pt,hypcap=false}
\captionof{figure}{Perplexity under quantization at different inference depths. Rows and columns correspond to quantization settings and evaluation corpora, respectively.}
\label{fig:stability_quantization}
\end{minipage}
\end{minipage}
\par\vspace{8pt}

\paragraph{Robustness under quantization.} Finally, we evaluate W8A16, W4A16, and W8A8 quantization at inference depths of 2, 4, and 8 loops. Figure~\ref{fig:stability_quantization} shows that RLT retains its perplexity advantage across all three quantization settings, inference budgets, and evaluation corpora. This extends the performance advantage to quantized computation, including reduced-loop inference. Quantization configurations and implementation details are provided in Appendix~\ref{app:stability_quantization}.

% ----------------------------------------------------------- %

\subsection{Latent Dynamics across Inference Depths}
\label{subsec:latent_dynamics}

We first compare SLT and RLT across inference depths, then use PCA visualizations to examine the latent dynamics underlying their performance differences. Both models are trained with eight loops and evaluated at smaller and larger inference budgets.

\begin{figure}[t]
\centering
\begin{minipage}[t][152pt][t]{0.36\linewidth}
\vspace{0pt}
\centering
\begingroup
\definecolor{rltgroup}{HTML}{F0F1F3}
\definecolor{rltrow}{HTML}{EBF3FA}
\definecolor{rltdivider}{HTML}{9298A0}
\setlength{\tabcolsep}{2pt}
\setlength{\aboverulesep}{1pt}
\setlength{\belowrulesep}{1pt}
\renewcommand{\arraystretch}{1.12}
\fontsize{8}{9.6}\selectfont
\scalebox{0.96}{%
\begin{tabular}{c!{\color{rltdivider}\vrule width .3pt}ccc!{\color{rltdivider}\vrule width .3pt}c}
\toprule
\multirow{2}{*}{\textbf{Model}} & \multicolumn{3}{c!{\color{rltdivider}\vrule width .3pt}}{\textbf{Perplexity} $\downarrow$} & \textbf{Acc (\%)} $\uparrow$ \\
\cmidrule(lr){2-4}\cmidrule(l){5-5}
 & \textbf{Pile} & \textbf{FWE} & \textbf{OWT} & \textbf{Avg Acc} \\
\midrule
\rowcolor{rltgroup}
\multicolumn{5}{c}{$\boldsymbol{K_{\mathrm{test}}=4}$} \\
\midrule
SLT & 13.01 & 33.62 & 30.87 & \textbf{42.95} \\
\rowcolor{rltrow}
RLT & \textbf{11.76} & \textbf{29.56} & \textbf{27.06} & 41.96 \\
\midrule
\rowcolor{rltgroup}
\multicolumn{5}{c}{$\boldsymbol{K_{\mathrm{test}}=2}$} \\
\midrule
SLT & 21.60 & 69.18 & 62.04 & 39.24 \\
\rowcolor{rltrow}
RLT & \textbf{13.85} & \textbf{35.82} & \textbf{32.81} & \textbf{40.76} \\
\bottomrule
\end{tabular}%
}
\endgroup
\vfill
\captionsetup{font=small,skip=4pt,justification=raggedright,singlelinecheck=false}
\captionof{table}{Fewer-loop inference with $K_{\mathrm{train}}=8$. \textbf{Bold} marks the better result at each depth; Avg Acc averages ten benchmarks.}
\label{tab:reduced-depth}
\end{minipage}\hfill
\begin{minipage}[t][152pt][t]{0.62\linewidth}
\vspace{0pt}
\centering
\makebox[\linewidth][r]{\includegraphics[width=1.03\linewidth]{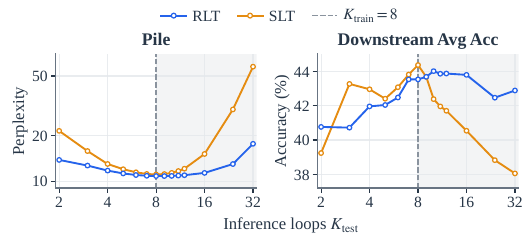}}
\vfill
\captionsetup{font=small,skip=4pt,justification=raggedright,singlelinecheck=false}
\captionof{figure}{Performance across inference depths for models trained with $K_{\mathrm{train}}=8$. The dashed line marks the training depth, and the shaded region indicates larger inference budgets.}
\label{fig:inference_depth}
\end{minipage}
\end{figure}

\paragraph{Adaptation to inference depth.} We first reduce the inference budget while keeping the trained weights fixed. Table~\ref{tab:reduced-depth} shows that RLT achieves lower perplexity on all three corpora at lower budgets. At $K_{\mathrm{test}}=2$, it reduces PPL by 35.9\%, 48.2\%, and 47.1\% on Pile, FWE, and OWT, respectively, while increasing average downstream accuracy from 39.24\% to 40.76\%. These results demonstrate that RLT better preserves language modeling quality when fewer recurrent steps are available.

We then extend the sweep to inference depths between 2 and 32 (Figure~\ref{fig:inference_depth}). The results show that RLT achieves lower Pile perplexity at every evaluated depth and exhibits a flatter performance profile around the training depth. Beyond eight loops, SLT's perplexity rises rapidly and its downstream accuracy declines, whereas RLT tolerates the increased depth more effectively. At $K_{\mathrm{test}}=32$, RLT achieves a PPL of 17.72 versus 57.72 for SLT and an average accuracy of 42.88\% versus 38.05\%. RLT thus supports a broader range of inference budgets, with improved performance retention both below and above the training depth. Evaluation details are provided in Appendix~\ref{app:inference_depth_evaluation}.

% Table and figure share a float so that they remain side by side.

\begin{figure}[!t]
\centering
\includegraphics[width=\linewidth]{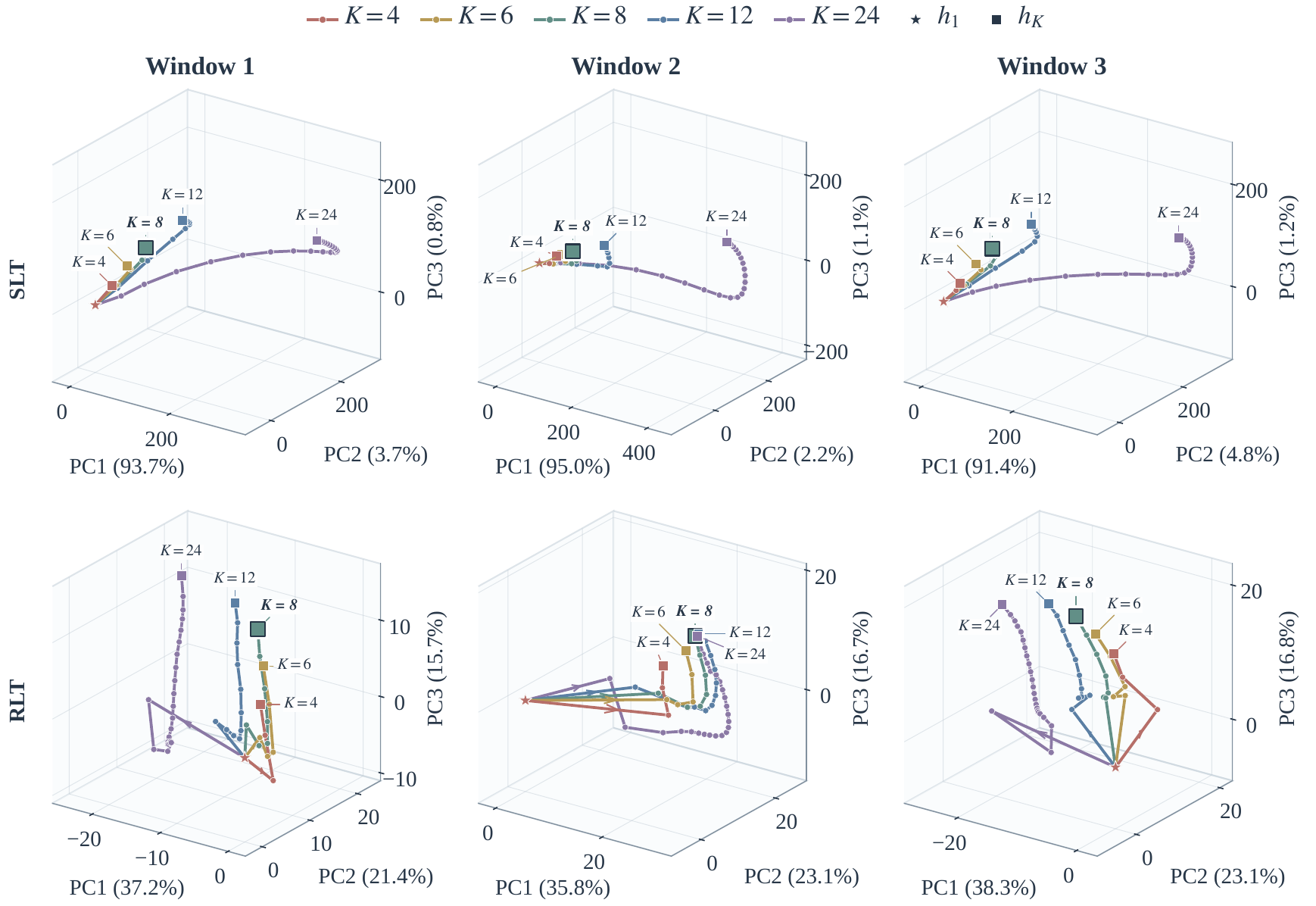}
\caption{3D PCA projections of final-token hidden outputs $\mathbf{h}$ for three OpenWebText windows. SLT and RLT are on the top and bottom rows, respectively. Colors indicate the inference budget; stars and squares mark $\mathbf{h}_1$ and $\mathbf{h}_K$. The outlined green square marks the matched-depth endpoint.}
\label{fig:latent_pca}
\end{figure}

\paragraph{Hidden-state trajectories.} To examine the latent dynamics behind these performance differences, we use 3D PCA to visualize the final token's hidden outputs $\mathbf{h}_1,\ldots,\mathbf{h}_K$ in three examples from randomly sampled OpenWebText windows (Figure~\ref{fig:latent_pca}). Each panel overlays inference budgets $K_{\mathrm{test}}\in\{4,6,8,12,24\}$ in a common projection, with the matched-depth endpoint at $K_{\mathrm{test}}=K_{\mathrm{train}}=8$ highlighted as a reference. Trajectory extraction, PCA construction, and visualizations of additional windows are provided in Appendix~\ref{app:latent_pca}.

Figure~\ref{fig:latent_pca} clearly shows that RLT dynamically adjusts its state-update step sizes using time information as the inference budget changes, while keeping its final hidden outputs concentrated around the matched-depth endpoint. Several RLT trajectories near the training depth also exhibit more uniform step-to-step changes than their SLT counterparts. In SLT, fewer loops fall short of the matched-depth endpoint, whereas additional loops drift beyond it and substantially expand the trajectory. For example, the terminal PC1 coordinate in Window 2 increases from 31.2 at $K_{\mathrm{test}}=4$ to 62.5 at $K_{\mathrm{test}}=8$ and 388.3 at $K_{\mathrm{test}}=24$, reaching over six times the matched-depth value. These contrasting behaviors agree with the schematic dynamics in Figure~\ref{fig:rlt-overview}(b) and connect RLT's greater tolerance of inference-depth changes with more controlled representation dynamics. They also align with RLT's remaining-time feedback schedule, which adjusts the amount of state advancement to the chosen loop budget.

\section{Conclusion}
\label{sec:conclusion}

We introduced Rectified Looped Transformer (RLT), which improves recurrent computation through a flow-inspired feedback rule without additional trainable feedback parameters or recurrent-block forward passes. RLT interpolates between the current input state and each loop's endpoint prediction according to the remaining time, separating prediction from state advancement. Our analysis establishes depth-uniform state and output bounds under time-dependent residual control and reduces worst-case initial-perturbation amplification bounds from exponential to polynomial in depth under a uniform Lipschitz condition. Experiments demonstrate consistently lower perplexity at matched depths, greater robustness to perturbations and inference-depth changes, and more controlled latent dynamics than direct feedback. Improved scale stability also supports more effective contrastive decoding and guidance. Our findings highlight state feedback as an effective design choice for jointly improving language modeling and the stability of latent recurrent computation.

\bibliography{ref}

@String(AAAI      = {Assoc. Adv. Artif. Intell.})

@String(EMNLP     = {Emp. Met. Nat. Lang. Proc.})

@String(AAAI      = {Association for the Advancement of Artificial Intelligence})

@String(EMNLP     = {Empirical Methods in Natural Language Processing})

@inproceedings{giannou2023,
  title     = {Looped {Transformers} as Programmable Computers},
  author    = {Giannou, Angeliki and Rajput, Shashank and Sohn, Jy-Yong
               and Lee, Kangwook and Lee, Jason D.
               and Papailiopoulos, Dimitris},
  booktitle = {Proceedings of the 40th International Conference on Machine Learning},
  series    = {Proceedings of Machine Learning Research},
  volume    = {202},
  pages     = {11398--11442},
  publisher = {PMLR},
  year      = {2023},
  url       = {https://proceedings.mlr.press/v202/giannou23a.html}
}

@article{movahedi2026,
  title   = {Fixed-Point Reasoners:
             Stable and Adaptive Deep Looped {Transformers}},
  author  = {Movahedi, Sajad and Milovanovi{\'c}, Vera
             and Feigin, Shlomo Libo and Theus, Alexander
             and Hofmann, Thomas and Boeva, Valentina
             and Rusch, T. Konstantin and Orvieto, Antonio},
  journal = {arXiv preprint arXiv:2606.18206},
  year    = {2026},
  url     = {https://arxiv.org/abs/2606.18206}
}

@article{cominetti2014,
  title   = {On the Rate of Convergence of
             {Krasnosel'ski{\u{\i}}--Mann} Iterations
             and Their Connection with Sums of {Bernoullis}},
  author  = {Cominetti, Roberto and Soto, Jos{\'e} A.
             and Vaisman, Jos{\'e}},
  journal = {Israel Journal of Mathematics},
  volume  = {199},
  number  = {2},
  pages   = {757--772},
  year    = {2014},
  doi     = {10.1007/s11856-013-0045-4}
}

@inproceedings{lipman2023,
  title     = {Flow Matching for Generative Modeling},
  author    = {Lipman, Yaron and Chen, Ricky T. Q.
               and Ben-Hamu, Heli and Nickel, Maximilian and Le, Matt},
  booktitle = {International Conference on Learning Representations},
  year      = {2023},
  url       = {https://openreview.net/forum?id=PqvMRDCJT9t}
}

@article{wang2026taps,
  title   = {Scheduling Recursive Reasoning in Looped {Transformers}},
  author  = {Wang, Boyuan and Yu, Chengyao and Ren, Jiaxi
             and Wei, Hongxin and Jing, Bingyi and Tao, Yuxin},
  journal = {arXiv preprint arXiv:2609.36653},
  year    = {2026},
  url     = {https://arxiv.org/abs/2609.36653}
}

@article{izermine2026,
  title   = {Unrolled Flow Models for Reasoning},
  author  = {Izermine, Faissal and Bai, Hanru and Davis, Oscar
             and Rusch, T. Konstantin},
  journal = {arXiv preprint arXiv:2610.09759},
  year    = {2026},
  url     = {https://arxiv.org/abs/2610.09759}
}

@inproceedings{saunshi2025reasoning,
  title     = {Reasoning with Latent Thoughts: On the Power of Looped Transformers},
  author    = {Saunshi, Nikunj and Dikkala, Nishanth and Li, Zhiyuan
               and Kumar, Sanjiv and Reddi, Sashank J.},
  booktitle = {The Thirteenth International Conference on Learning Representations},
  year      = {2025},
  url       = {https://openreview.net/forum?id=din0lGfZFd}
}

@inproceedings{geiping2025scaling,
  title     = {Scaling up Test-Time Compute with Latent Reasoning:
               A Recurrent Depth Approach},
  author    = {Geiping, Jonas and McLeish, Sean and Jain, Neel
               and Kirchenbauer, John and Singh, Siddharth
               and Bartoldson, Brian and Kailkhura, Bhavya
               and Bhatele, Abhinav and Goldstein, Tom},
  booktitle = {Advances in Neural Information Processing Systems},
  volume    = {38},
  pages     = {41340--41391},
  year      = {2025},
  doi       = {10.52202/085713-1380},
  url       = {https://proceedings.neurips.cc/paper_files/paper/2025/hash/3b01972cf31e6fa0fe29e4b8b5c2a0a1-Abstract-Conference.html}
}

@article{zhu2025scaling,
  title   = {Scaling Latent Reasoning via Looped Language Models},
  author  = {Zhu, Rui-Jie and Wang, Zixuan and Hua, Kai
             and Zhang, Tianyu and Li, Ziniu and Que, Haoran
             and Wei, Boyi and Wen, Zixin and Yin, Fan
             and Xing, He and Li, Lu and Shi, Jiajun
             and Ma, Kaijing and Li, Shanda and Kergan, Taylor
             and Smith, Andrew and Qu, Xingwei and Hui, Mude
             and Wu, Bohong and Min, Qiyang and Huang, Hongzhi
             and Zhou, Xun and Ye, Wei and Liu, Jiaheng
             and Yang, Jian and Shi, Yunfeng and Lin, Chenghua
             and Zhao, Enduo and Cai, Tianle and Zhang, Ge
             and Huang, Wenhao and Bengio, Yoshua and Eshraghian, Jason},
  journal = {arXiv preprint arXiv:2510.25741},
  year    = {2025},
  url     = {https://arxiv.org/abs/2510.25741}
}

@inproceedings{cho2014learning,
  title     = {Learning Phrase Representations using {RNN} Encoder{--}Decoder
               for Statistical Machine Translation},
  author    = {Cho, Kyunghyun and van Merri{\"e}nboer, Bart
               and Gulcehre, Caglar and Bahdanau, Dzmitry
               and Bougares, Fethi and Schwenk, Holger and Bengio, Yoshua},
  booktitle = {Proceedings of the 2014 Conference on Empirical Methods
               in Natural Language Processing ({EMNLP})},
  pages     = {1724--1734},
  year      = {2014},
  publisher = {Association for Computational Linguistics},
  doi       = {10.3115/v1/D14-1179},
  url       = {https://aclanthology.org/D14-1179/}
}

@article{chung2014empirical,
  title   = {Empirical Evaluation of Gated Recurrent Neural Networks
             on Sequence Modeling},
  author  = {Chung, Junyoung and Gulcehre, Caglar
             and Cho, KyungHyun and Bengio, Yoshua},
  journal = {arXiv preprint arXiv:1412.3555},
  year    = {2014},
  url     = {https://arxiv.org/abs/1412.3555}
}

@article{suleymanzade2026thinking,
  title   = {Thinking with Looped Flows},
  author  = {Suleymanzade, Ayhan and Lee, Chanhyuk
             and Eijkelboom, Floor and Boffi, Nicholas M.
             and Ceylan, {\.{I}}smail {\.{I}}lkan and Kim, Jinwoo},
  journal = {arXiv preprint arXiv:2609.11801},
  year    = {2026},
  url     = {https://arxiv.org/abs/2609.11801}
}

@article{gao2026loop,
  title   = {Loop the {Loopies}!},
  author  = {Gao, Zitian and Chen, Yilong and Xiao, Yihao
             and Yang, Xinyu and Tao, Ran and Zhou, Joey and Dai, Bryan},
  journal = {arXiv preprint arXiv:2607.16051},
  year    = {2026},
  url     = {https://arxiv.org/abs/2607.16051}
}

@inproceedings{jeddi2026loopformer,
  title     = {{LoopFormer}: Elastic-Depth Looped Transformers
               for Latent Reasoning via Shortcut Modulation},
  author    = {Jeddi, Ahmadreza and Ciccone, Marco and Taati, Babak},
  booktitle = {The Fourteenth International Conference on Learning Representations},
  year      = {2026},
  url       = {https://openreview.net/forum?id=RzYXb5YWBs}
}

@inproceedings{yang2026stabilizing,
  title     = {Stabilizing Recurrent Dynamics for Test-Time Scalable
               Latent Reasoning in Looped Language Models},
  author    = {Yang, Xiao-Wen and Han, Ziyu and Zhang, Xi-Hua
               and Wei, Wen-Da and Shao, Jie-Jing
               and Guo, Lan-Zhe and Li, Yu-Feng},
  booktitle = {Proceedings of the 43rd International Conference on Machine Learning},
  series    = {Proceedings of Machine Learning Research},
  volume    = {306},
  pages     = {144670--144682},
  year      = {2026},
  publisher = {PMLR},
  url       = {https://proceedings.mlr.press/v306/yang26t.html}
}

@inproceedings{liu2023flow,
  title     = {Flow Straight and Fast: Learning to Generate and Transfer
               Data with Rectified Flow},
  author    = {Liu, Xingchao and Gong, Chengyue and Liu, Qiang},
  booktitle = {The Eleventh International Conference on Learning Representations},
  year      = {2023},
  url       = {https://openreview.net/forum?id=XVjTT1nw5z}
}

@inproceedings{dehghani2019universal,
  title     = {Universal {Transformers}},
  author    = {Dehghani, Mostafa and Gouws, Stephan and Vinyals, Oriol
               and Uszkoreit, Jakob and Kaiser, {\L}ukasz},
  booktitle = {International Conference on Learning Representations},
  year      = {2019},
  url       = {https://openreview.net/forum?id=HyzdRiR9Y7}
}

@inproceedings{lan2020albert,
  title     = {{ALBERT}: A Lite {BERT} for Self-supervised Learning
               of Language Representations},
  author    = {Lan, Zhenzhong and Chen, Mingda and Goodman, Sebastian
               and Gimpel, Kevin and Sharma, Piyush and Soricut, Radu},
  booktitle = {International Conference on Learning Representations},
  year      = {2020},
  url       = {https://openreview.net/forum?id=H1eA7AEtvS}
}

@inproceedings{bae2025relaxed,
  title     = {Relaxed Recursive {Transformers}: Effective Parameter Sharing
               with Layer-wise {LoRA}},
  author    = {Bae, Sangmin and Fisch, Adam and Harutyunyan, Hrayr
               and Ji, Ziwei and Kim, Seungyeon and Schuster, Tal},
  booktitle = {International Conference on Learning Representations},
  year      = {2025},
  url       = {https://openreview.net/forum?id=WwpYSOkkCt}
}

@inproceedings{bae2025mixture,
  title     = {Mixture-of-Recursions: Learning Dynamic Recursive Depths
               for Adaptive Token-Level Computation},
  author    = {Bae, Sangmin and Kim, Yujin and Bayat, Reza and Kim, Sungnyun
               and Ha, Jiyoun and Schuster, Tal and Fisch, Adam
               and Harutyunyan, Hrayr and Ji, Ziwei and Courville, Aaron
               and Yun, Se-Young},
  booktitle = {Advances in Neural Information Processing Systems},
  volume    = {38},
  pages     = {96572--96617},
  year      = {2025},
  doi       = {10.52202/085713-3229},
  url       = {https://proceedings.neurips.cc/paper_files/paper/2025/hash/8b08bbf8b420faa6eeb4020720582ec7-Abstract-Conference.html}
}

@inproceedings{fu2026think,
  title     = {Think-at-Hard: Dynamic Looped {Transformers} for Improved Reasoning},
  author    = {Fu, Tianyu and You, Yichen and Chen, Zekai and Dai, Guohao
               and Yang, Huazhong and Wang, Yu},
  booktitle = {Proceedings of the 43rd International Conference on Machine Learning},
  series    = {Proceedings of Machine Learning Research},
  volume    = {306},
  pages     = {32129--32154},
  publisher = {PMLR},
  year      = {2026},
  url       = {https://proceedings.mlr.press/v306/fu26t.html}
}

@article{you2026improving,
  title   = {Improving Test-Time Scaling with Adaptive Looped {Transformers}},
  author  = {You, Yichen and Fu, Tianyu and Feng, Aosong and Lv, Xingtai
             and Ning, Xuefei and Ding, Ning and Wang, Yu},
  journal = {arXiv preprint arXiv:2609.35748},
  year    = {2026},
  url     = {https://arxiv.org/abs/2609.35748}
}

@article{chen2026loopmoe,
  title   = {{LoopMoE}: Unifying Iterative Computation with Mixture-of-Experts
             for Language Modeling},
  author  = {Chen, Wenkai and Li, Tianshu and Huang, Wenyong and Yin, Yichun
             and Shang, Lifeng and Qin, Chengwei},
  journal = {arXiv preprint arXiv:2606.04438},
  year    = {2026},
  url     = {https://arxiv.org/abs/2606.04438}
}

@article{lee2026sparse,
  title   = {Sparse Layers are Critical to Scaling Looped Language Models},
  author  = {Lee, Ryan and Biloki, Jacob and Hu, Edward J. and May, Jonathan},
  journal = {arXiv preprint arXiv:2605.09165},
  year    = {2026},
  url     = {https://arxiv.org/abs/2605.09165}
}

@article{ladhak2026one,
  title   = {One Model to Loop Them All: Elastic Latent Recurrence
             for Language Models},
  author  = {Ladhak, Faisal and Acharya, Shantanu and Serano, Santiago Akle
             and Sun, Simeng and Ginsburg, Boris and Puvvada, Krishna C.},
  journal = {arXiv preprint arXiv:2610.12582},
  year    = {2026},
  url     = {https://arxiv.org/abs/2610.12582}
}

@article{gao2020pile,
  title   = {The {Pile}: An {800GB} Dataset of Diverse Text for Language Modeling},
  author  = {Gao, Leo and Biderman, Stella and Black, Sid and Golding, Laurence
             and Hoppe, Travis and Foster, Charles and Phang, Jason and He, Horace
             and Thite, Anish and Nabeshima, Noa and Presser, Shawn and Leahy, Connor},
  journal = {arXiv preprint arXiv:2101.00027},
  year    = {2020},
  url     = {https://arxiv.org/abs/2101.00027}
}

@inproceedings{penedo2024fineweb,
  title     = {The {FineWeb} Datasets: Decanting the Web for the Finest Text Data at Scale},
  author    = {Penedo, Guilherme and Kydl{\'i}{\v{c}}ek, Hynek and Ben Allal, Loubna
               and Lozhkov, Anton and Mitchell, Margaret and Raffel, Colin
               and von Werra, Leandro and Wolf, Thomas},
  booktitle = {Advances in Neural Information Processing Systems},
  volume    = {37},
  pages     = {30811--30849},
  publisher = {Curran Associates, Inc.},
  year      = {2024},
  doi       = {10.52202/079017-0970},
  url       = {https://proceedings.neurips.cc/paper_files/paper/2024/hash/370df50ccfdf8bde18f8f9c2d9151bda-Abstract-Datasets_and_Benchmarks_Track.html}
}

@misc{gokaslan2019openwebtext,
  title        = {{OpenWebText} Corpus},
  author       = {Gokaslan, Aaron and Cohen, Vanya},
  year         = {2019},
  url = {https://skylion007.github.io/OpenWebTextCorpus/}
}

@misc{gao2023framework,
  title     = {A Framework for Few-Shot Language Model Evaluation},
  author    = {Gao, Leo and Tow, Jonathan and Abbasi, Baber and Biderman, Stella
               and Black, Sid and DiPofi, Anthony and Foster, Charles
               and Golding, Laurence and Hsu, Jeffrey and Le Noac'h, Alain
               and Li, Haonan and McDonell, Kyle and Muennighoff, Niklas
               and Ociepa, Chris and Phang, Jason and Reynolds, Laria
               and Schoelkopf, Hailey and Skowron, Aviya and Sutawika, Lintang
               and Tang, Eric and Thite, Anish and Wang, Ben and Wang, Kevin
               and Zou, Andy},
  year      = {2023},
  month     = dec,
  publisher = {Zenodo},
  version   = {v0.4.0},
  doi       = {10.5281/zenodo.10256836},
  url       = {https://zenodo.org/records/10256836}
}

@inproceedings{roemmele2011choice,
  title     = {Choice of Plausible Alternatives: An Evaluation of Commonsense Causal Reasoning},
  author    = {Roemmele, Melissa and Bejan, Cosmin Adrian and Gordon, Andrew S.},
  booktitle = {AAAI Spring Symposium on Logical Formalizations of Commonsense Reasoning},
  year      = {2011},
  url       = {https://commonsensereasoning.org/2011/papers/Roemmele.pdf}
}

@inproceedings{zellers2019hellaswag,
  title     = {{HellaSwag}: Can a Machine Really Finish Your Sentence?},
  author    = {Zellers, Rowan and Holtzman, Ari and Bisk, Yonatan and Farhadi, Ali and Choi, Yejin},
  booktitle = {Proceedings of the 57th Annual Meeting of the Association for Computational Linguistics},
  pages     = {4791--4800},
  publisher = {Association for Computational Linguistics},
  year      = {2019},
  doi       = {10.18653/v1/P19-1472},
  url       = {https://aclanthology.org/P19-1472/}
}

@inproceedings{paperno2016lambada,
  title     = {The {LAMBADA} Dataset: Word Prediction Requiring a Broad Discourse Context},
  author    = {Paperno, Denis and Kruszewski, Germ{\'a}n and Lazaridou, Angeliki
               and Pham, Ngoc Quan and Bernardi, Raffaella and Pezzelle, Sandro
               and Baroni, Marco and Boleda, Gemma and Fern{\'a}ndez, Raquel},
  booktitle = {Proceedings of the 54th Annual Meeting of the Association for Computational Linguistics (Volume 1: Long Papers)},
  pages     = {1525--1534},
  publisher = {Association for Computational Linguistics},
  year      = {2016},
  doi       = {10.18653/v1/P16-1144},
  url       = {https://aclanthology.org/P16-1144/}
}

@inproceedings{mihaylov2018suit,
  title     = {Can a Suit of Armor Conduct Electricity? A New Dataset for Open Book Question Answering},
  author    = {Mihaylov, Todor and Clark, Peter and Khot, Tushar and Sabharwal, Ashish},
  booktitle = {Proceedings of the 2018 Conference on Empirical Methods in Natural Language Processing},
  pages     = {2381--2391},
  publisher = {Association for Computational Linguistics},
  year      = {2018},
  doi       = {10.18653/v1/D18-1260},
  url       = {https://aclanthology.org/D18-1260/}
}

@inproceedings{bisk2020piqa,
  title     = {{PIQA}: Reasoning about Physical Commonsense in Natural Language},
  author    = {Bisk, Yonatan and Zellers, Rowan and Le Bras, Ronan and Gao, Jianfeng and Choi, Yejin},
  booktitle = {Proceedings of the AAAI Conference on Artificial Intelligence},
  volume    = {34},
  pages     = {7432--7439},
  year      = {2020},
  doi       = {10.1609/aaai.v34i05.6239},
  url       = {https://ojs.aaai.org/index.php/AAAI/article/view/6239}
}

@inproceedings{lai2017race,
  title     = {{RACE}: Large-scale {ReAding} Comprehension Dataset From Examinations},
  author    = {Lai, Guokun and Xie, Qizhe and Liu, Hanxiao and Yang, Yiming and Hovy, Eduard},
  booktitle = {Proceedings of the 2017 Conference on Empirical Methods in Natural Language Processing},
  pages     = {785--794},
  publisher = {Association for Computational Linguistics},
  year      = {2017},
  doi       = {10.18653/v1/D17-1082},
  url       = {https://aclanthology.org/D17-1082/}
}

@inproceedings{welbl2017crowdsourcing,
  title     = {Crowdsourcing Multiple Choice Science Questions},
  author    = {Welbl, Johannes and Liu, Nelson F. and Gardner, Matt},
  booktitle = {Proceedings of the 3rd Workshop on Noisy User-generated Text},
  pages     = {94--106},
  publisher = {Association for Computational Linguistics},
  year      = {2017},
  doi       = {10.18653/v1/W17-4413},
  url       = {https://aclanthology.org/W17-4413/}
}

@article{clark2018think,
  title   = {Think You Have Solved Question Answering? Try {ARC}, the {AI2} Reasoning Challenge},
  author  = {Clark, Peter and Cowhey, Isaac and Etzioni, Oren and Khot, Tushar
             and Sabharwal, Ashish and Schoenick, Carissa and Tafjord, Oyvind},
  journal = {arXiv preprint arXiv:1803.05457},
  year    = {2018},
  url     = {https://arxiv.org/abs/1803.05457}
}

@inproceedings{sap2019social,
  title     = {Social {IQA}: Commonsense Reasoning about Social Interactions},
  author    = {Sap, Maarten and Rashkin, Hannah and Chen, Derek and Le Bras, Ronan and Choi, Yejin},
  booktitle = {Proceedings of the 2019 Conference on Empirical Methods in Natural Language Processing and the 9th International Joint Conference on Natural Language Processing (EMNLP-IJCNLP)},
  pages     = {4463--4473},
  publisher = {Association for Computational Linguistics},
  year      = {2019},
  doi       = {10.18653/v1/D19-1454},
  url       = {https://aclanthology.org/D19-1454/}
}

@article{sakaguchi2021winogrande,
  title     = {{WinoGrande}: An Adversarial Winograd Schema Challenge at Scale},
  author    = {Sakaguchi, Keisuke and Le Bras, Ronan and Bhagavatula, Chandra and Choi, Yejin},
  journal   = {Communications of the ACM},
  volume    = {64},
  number    = {9},
  pages     = {99--106},
  publisher = {Association for Computing Machinery},
  year      = {2021},
  doi       = {10.1145/3474381},
  url       = {https://doi.org/10.1145/3474381}
}

@inproceedings{peebles2023scalable,
  title     = {Scalable Diffusion Models with Transformers},
  author    = {Peebles, William and Xie, Saining},
  booktitle = {Proceedings of the IEEE/CVF International Conference on Computer Vision},
  pages     = {4195--4205},
  year      = {2023},
  url       = {https://openaccess.thecvf.com/content/ICCV2023/html/Peebles_Scalable_Diffusion_Models_with_Transformers_ICCV_2023_paper.html}
}

@inproceedings{li2023contrastive,
  title     = {Contrastive Decoding: Open-ended Text Generation as Optimization},
  author    = {Li, Xiang Lisa and Holtzman, Ari and Fried, Daniel and Liang, Percy and Eisner, Jason and Hashimoto, Tatsunori and Zettlemoyer, Luke and Lewis, Mike},
  booktitle = {Proceedings of the 61st Annual Meeting of the Association for Computational Linguistics (Volume 1: Long Papers)},
  pages     = {12286--12312},
  publisher = {Association for Computational Linguistics},
  year      = {2023},
  doi       = {10.18653/v1/2023.acl-long.687},
  url       = {https://aclanthology.org/2023.acl-long.687/}
}

@inproceedings{yu2026loopcd,
  title     = {{LoopCD}: Loop-wise Contrastive Decoding for Improving Reasoning in Looped Language Models},
  author    = {Yu, Byeongho and So, Junhyuk and Park, Eunhyeok},
  booktitle = {Proceedings of the 2026 Conference on Empirical Methods in Natural Language Processing},
  pages     = {31185--31199},
  publisher = {Association for Computational Linguistics},
  year      = {2026},
  url       = {https://aclanthology.org/2026.emnlp-main.1367/}
}

@article{liu2026decoding,
  title   = {Decoding Looped {Transformers} Better for (Almost) Free},
  author  = {Liu, Weihao and Zheng, Huangjie and Chen, Tianrong and Dilip, Rohit and Bai, Richard He and Jiao, Yizhu and Wang, Yuyang and Zhang, Ruixiang},
  journal = {arXiv preprint arXiv:2610.02185},
  year    = {2026},
  url     = {https://arxiv.org/abs/2610.02185}
}

@inproceedings{karras2024guiding,
  title     = {Guiding a Diffusion Model with a Bad Version of Itself},
  author    = {Karras, Tero and Aittala, Miika and Kynk{\"a}{\"a}nniemi, Tuomas and Lehtinen, Jaakko and Aila, Timo and Laine, Samuli},
  booktitle = {Advances in Neural Information Processing Systems},
  volume    = {37},
  pages     = {52996--53021},
  publisher = {Curran Associates, Inc.},
  year      = {2024},
  doi       = {10.52202/079017-1679},
  url       = {https://proceedings.neurips.cc/paper_files/paper/2024/hash/5ee7ed60a7e8169012224dec5fe0d27f-Abstract-Conference.html}
}
\bibliographystyle{iclr2027_conference_nourl}

\newpage
\appendix

\section{Proofs of Theorems and Propositions}
\label{app:proofs}

\subsection{Proof of Theorem~\ref{thm:bounded} and Corollaries~\ref{cor:directstate}--\ref{cor:radial}}
\label{app:proof_of_thm1}

\begin{proof}
Applying the triangle inequality gives
\begin{align}
\norm{\mathbf{x}_{k+1}}\le\norm{\mathbf{x}_k}
+\eta_k\norm{\mathbf{h}_{k+1}-\mathbf{x}_k}
\le\norm{\mathbf{x}_k}+M\eta_k(1-t_k)^\alpha.
\label{eq:triangle}
\end{align}
Summing Eq.~(\ref{eq:triangle}) with respect to $k$ proves Eq.~(\ref{eq:generalbound}). Our schedule (Eq.~(\ref{eq:schedule})) gives
\begin{align}
\norm{\mathbf{x}_k}\le\norm{\mathbf{x}_0}
+M\Delta t\sum_{j=0}^{k-1}(1-t_j)^{\alpha-1}
\le\norm{\mathbf{x}_0}+\frac M\alpha\bigl[1-(1-t_k)^\alpha\bigr].
\end{align}
For $k<K$, this further yields
\begin{align}
\norm{\mathbf{h}_{k+1}}
\le\norm{\mathbf{x}_k}+M(1-t_k)^\alpha
\le\norm{\mathbf{x}_0}+\frac M\alpha
-M\left(\frac1\alpha-1\right)(1-t_k)^\alpha
\le\norm{\mathbf{x}_0}+\frac M\alpha.
\end{align}
Thus Eq.~(\ref{eq:bothbound}) follows.

Setting $\eta_k\equiv1$ for direct feedback in Eq.~(\ref{eq:generalbound}) gives Eq.~(\ref{eq:directstate}). We further note that the bound in Eq.~(\ref{eq:directstate}) can be attained. Specifically, when $\mathbf{x}_0=0$ and $\mathbf{F}_\theta(\mathbf{x},t)=\mathbf{x}+M(1-t)^\alpha \mathbf{u}$, where $\norm{\mathbf{u}}=1$, then for $M>0$,
\begin{align}
\norm{\mathbf{x}_K}=MK^{-\alpha}\sum_{j=1}^Kj^\alpha
\sim\frac{MK}{\alpha+1}.
\end{align}

Finally, under the radial condition in Corollary~\ref{cor:radial}, we have
\begin{align}
\norm{\mathbf{x}_{k+1}}^2-\norm{\mathbf{x}_k}^2
=2\eta_k\ip{\mathbf{x}_k}{\mathbf{h}_{k+1}-\mathbf{x}_k}
+\eta_k^2\norm{\mathbf{h}_{k+1}-\mathbf{x}_k}^2
\le M^2\eta_k^2(1-t_k)^{2\alpha}.
\end{align}
Summing and noting that $j=K-k$,
\begin{align}
\max_{0\le k\le K}\norm{\mathbf{x}_k}^2
\le\norm{\mathbf{x}_0}^2+M^2\sum_{k=0}^{K-1}\eta_k^2(1-t_k)^{2\alpha}
=\norm{\mathbf{x}_0}^2+M^2K^{-2\alpha}\sum_{j=1}^Kj^{2\alpha-2}.
\end{align}
The additive term is
$\mathcal{O}(K^{-2\alpha})$ for $0<\alpha<1/2$,
$\mathcal{O}((\log K)/K)$ for $\alpha=1/2$, and $\mathcal{O}(K^{-1})$ for $1/2<\alpha\le1$.
At $\alpha=1$, the sum equals $K$, giving a vanishing bound as $\mathcal{O}(M^2/K)$.
\end{proof}

\subsection{Proof of Theorem~\ref{thm:perturbation} and Corollary~\ref{cor:amplification}}
\label{app:proof_of_thm2}

\begin{proof}
For two trajectories with the same coefficients,
\begin{align}
\norm{\mathbf{x}_{k+1}-\widetilde{\mathbf{x}}_{k+1}}
&\le(1-\eta_k)\norm{\mathbf{x}_k-\widetilde{\mathbf{x}}_k}
+\eta_k\norm{\mathbf{F}_\theta(\mathbf{x}_k,t_k)-\mathbf{F}_\theta(\widetilde{\mathbf{x}}_k,t_k)}\nonumber
\\
&\le(1-\eta_k+\eta_kL_k)\norm{\mathbf{x}_k-\widetilde{\mathbf{x}}_k}.
\end{align}
Eq.~(\ref{eq:product}) follows by taking the product over $k$. Setting $\eta_k\equiv1$ gives the direct-feedback bounds.

For our schedule (Eq.~(\ref{eq:schedule})) and $L_k\le L$,
\begin{align}
\prod_{k=0}^{K-1}(1-\eta_k+\eta_kL_k)
\le\prod_{j=1}^K\left(1+\frac{L-1}{j}\right).
\end{align}
Note that $\log(1+a)\le a$, $\sum_{j=1}^Kj^{-1}\le1+\log K$ and $L > 1$, thus
\[
\prod_{j=1}^K\left(1+\frac{L-1}{j}\right)
\le\exp\left((L-1)\sum_{j=1}^K\frac1j\right)
\le e^{L-1}K^{L-1}.
\]
This proves Eq.~(\ref{eq:polynomial}).
\end{proof}

\subsection{Proof of Proposition~\ref{prop:fixedpoint}}
\label{app:proof_of_prop3}

\begin{proof}
Using the result of the Krasnosel'skii--Mann method in~\citep[Proposition~11]{cominetti2014} with the infimum taken over all fixed points, we have
\begin{align}
\norm{\mathbf{F}_\theta(\mathbf{x}_K)-\mathbf{x}_K}
\le
\frac{\dist(\mathbf{x}_0,\Fix(\mathbf{F}_\theta))}
{\sqrt{\sum_{k=0}^{K-1}\eta_k(1-\eta_k)}}.
\end{align}
Here $\dist(\mathbf{x}_0,\Fix(\mathbf{F}_\theta))=\inf_{\mathbf{x}^\star\in\Fix(\mathbf{F}_\theta)}\norm{\mathbf{x}_0-\mathbf{x}^\star}$.
For schedule in Eq.~(\ref{eq:schedule}) and $K\ge2$,
\begin{align}
\sum_{k=0}^{K-1}\eta_k(1-\eta_k)
=\sum_{j=2}^K\frac{j-1}{j^2}
\ge\frac12\sum_{j=2}^K\frac1j
\ge\frac12\log\frac{K+1}{2}
\ge\frac14\log K.
\end{align}
Substitution proves Eq.~(\ref{eq:fixedresidual}).
\end{proof}

\section{Local Spectral Analysis}
\label{app:spectral}

Here we further state the local stability consequence for interpolated feedback in Eq.~(\ref{eq:update}) following the damping analysis of \citet[Appendix~A.3]{movahedi2026}. 

\begin{appproposition}
\label{prop:spectral}
Fix $t$, and suppose $\mathbf{F}_\theta(\cdot,t)$ is continuously differentiable near a fixed point $\mathbf{x}^\star$. If every eigenvalue $\lambda$ of $\partial_\mathbf{x}\mathbf{F}_\theta(\mathbf{x}^\star,t)$ satisfies $\operatorname{Re}\lambda<1$, then there exists $\eta\in(0,1]$ such that the iteration
\begin{align}
\mathbf{x}_{k+1}=(1-\eta)\mathbf{x}_k+\eta \mathbf{F}_\theta(\mathbf{x}_k,t)
\end{align}
is locally asymptotically stable at $\mathbf{x}^\star$.
\end{appproposition}
\begin{proof}
By the eigenvalue mapping in \citet[Appendix~A.3]{movahedi2026}, interpolation maps each eigenvalue $\lambda$ of $\partial_\mathbf{x}\mathbf{F}_\theta(\mathbf{x}^\star,t)$ to $\mu=1+\eta(\lambda-1)$.
Since all real parts of $\lambda$ are below one, we can always choose
\begin{align}
0<\eta<
\min_{\lambda}\frac{2(1-\operatorname{Re}\lambda)}{|\lambda-1|^2}.
\end{align}
Thus all mapped eigenvalues satisfy $|\mu|<1$, yielding local asymptotic stability.
\end{proof}
Proposition~\ref{prop:spectral} shows that interpolation can stabilize a fixed point even when the original Jacobian has eigenvalues outside the unit disk. The interpolation coefficient controls how far each eigenvalue moves from $1$ toward its original value, allowing local stability whenever all real parts lie below one.

% ----------------------------------------------------------- %

\section{Model Architecture}
\label{app:model_architecture}

As illustrated in Figure~\ref{fig:rlt-overview}(a), we adopt the LoopFormer~\citep{jeddi2026loopformer} architecture, with token embeddings and learned absolute positional embeddings forming the initial state $\mathbf{x}_0$. The shared stack contains three Transformer layers, each comprising causal multi-head self-attention (MHSA) and a two-layer feed-forward network (FFN) with GELU activations. Time and step-size conditioning uses AdaLN-style modulation~\citep{peebles2023scalable} with RMSNorm: the conditioning vector controls the scales of normalized features and the gates on the attention and feed-forward residual branches.

\begin{table}[htbp]
    \centering
    \caption{Default model configuration.}
    \label{tab:rlt_model_configuration}
    \small
    \begin{tblr}{
        width = 0.99\linewidth,
        colspec = {Q[l]Q[l]},
        row{1} = {font=\bfseries},
        colsep = 30pt,
        rowsep = 1pt,
        column{1} = {leftsep=10pt},
        column{Z} = {rightsep=10pt},
    }
        \toprule
        Configuration & Setting \\
        \midrule
        Number of shared Transformer layers & 3 \\
        Default training / inference loops $(K_{\mathrm{train}}/K_{\mathrm{test}})$ & 8 / 8 \\
        Hidden dimension $d$ & 2,048 \\
        Attention heads / head dimension & 32 / 64 \\
        Feed-forward intermediate dimension & 5,120 \\
        Feed-forward activation & GELU \\
        Maximum context length & 1,024 \\
        Tokenizer & GPT-2 BPE \\
        Positional encoding & Learned absolute embeddings \\
        Dropout & 0 \\
        RMSNorm $\epsilon$ & $2^{-23}$ \\
        Physical parameters & 278M \\
        \bottomrule
    \end{tblr}
\end{table}

\paragraph{Time and step-size embeddings.}
We encode time and step size with sinusoidal features. For a scalar $\tau$, define
\begin{equation}
    \omega_j=10000^{-2j/d_{\mathrm f}},
    \qquad j=0,\ldots,d_{\mathrm f}/2-1,
    \qquad d_{\mathrm f}=256,
    \label{eq:app_time_frequencies}
\end{equation}
\begin{equation}
    \mathbf{s}(\tau)=\operatorname{concat}\!\left(
        \bigl(\cos(\tau\omega_j)\bigr)_{j=0}^{d_{\mathrm f}/2-1},
        \bigl(\sin(\tau\omega_j)\bigr)_{j=0}^{d_{\mathrm f}/2-1}
    \right).
    \label{eq:app_time_features}
\end{equation}
Two separate MLPs then produce the conditioning vector:
\begin{equation}
    \mathbf{c}_k=
    \operatorname{MLP}_{t}\!\left(\mathbf{s}(t_k)\right)
    +\operatorname{MLP}_{\Delta t}\!\left(\mathbf{s}(\Delta t)\right)
    \in\mathbb R^d.
    \label{eq:app_time_conditioning}
\end{equation}
Each MLP has dimensions $256\rightarrow2048\rightarrow2048$, with a SiLU activation between linear layers. The conditioning vector is shared across token positions and supplied to each layer of the recurrent stack.

\paragraph{Recurrent computation and feedback.}
Consistent with Section~\ref{subsec:looped_computation_and_state_feedback}, let $K$ denote the loop count for the current computation, taking $K=K_{\mathrm{train}}$ during training and $K=K_{\mathrm{test}}$ during inference. On the uniform grid in Eq.~(\ref{eq:uniform_time_grid}), the recurrent computation is
\begin{equation}
    \begin{aligned}
        \mathbf{h}_{k+1} &= \mathbf{F}_\theta(\mathbf{x}_k,t_k), \\
        \mathbf{x}_{k+1} &= (1-\eta_k)\mathbf{x}_k+\eta_k \mathbf{h}_{k+1},
        \qquad k=0,\ldots,K-1,
    \end{aligned}
    \label{eq:app_recurrent_update}
\end{equation}
with feedback coefficients
\begin{equation}
    \eta_k=
    \begin{cases}
        1, & \mathrm{SLT}, \\[2mm]
        \dfrac{\Delta t}{1-t_k}=\dfrac{1}{K-k}, & \mathrm{RLT}.
    \end{cases}
    \label{eq:app_feedback_coefficients}
\end{equation}
Changing $K_{\mathrm{test}}$ updates the time grid, step-size conditioning, and RLT feedback coefficients accordingly.

Each hidden output is decoded using the same readout:
\begin{equation}
    \mathcal D(\mathbf{h})=\mathbf{W}\,\operatorname{RMSNorm}(\mathbf{h}).
    \label{eq:app_shared_readout}
\end{equation}
Here, $\mathbf{W}$ is the weight matrix of \texttt{lm\_head}, shared with the input token embedding matrix.

\section{Training and Implementation Details}
\label{app:training_implementation}

SLT and RLT use the same initial weights, training examples, and data order. Each model is trained with a fixed $K_{\mathrm{train}}$ and uniform all-step next-token prediction supervision. Using the per-loop loss defined in Eq.~(\ref{eq:readout_supervision}), the training objective is
\begin{equation}
    \ell_k=\operatorname{CE}\!\left(\mathcal D(\mathbf{h}_k),\mathbf{y}\right),
    \qquad
    \mathcal L=\frac{1}{K_{\mathrm{train}}}
    \sum_{k=1}^{K_{\mathrm{train}}}\ell_k.
    \label{eq:app_training_objective}
\end{equation}
Cross-entropy is averaged over tokens and sequences within each batch.

We use AdamW with linear learning-rate warmup followed by cosine decay. Weight decay is applied to matrix-valued parameters, excluding biases and normalization parameters.

\begin{table}[htbp]
    \centering
    \caption{Training hyperparameters shared by SLT and RLT.}
    \label{tab:rlt_training_hyperparameters}
    \small
    \begin{tblr}{
        width = 0.90\linewidth,
        colspec = {Q[l]Q[c]},
        row{1} = {font=\bfseries},
        colsep = 40pt,
        rowsep = 1pt,
        column{1} = {leftsep=15pt},
        column{Z} = {rightsep=15pt},
    }
        \toprule
        Hyperparameter & Setting \\
        \midrule
        Optimizer & AdamW \\
        Peak learning rate & $6\times10^{-4}$ \\
        Minimum learning rate & $6\times10^{-5}$ \\
        Warmup steps & 4,000 \\
        Optimizer steps & 14,000 \\
        Adam $(\beta_1,\beta_2)$ & $(0.9,0.95)$ \\
        Weight decay & 0.2 \\
        Gradient norm clipping & 1.0 \\
        Global Batch Size & 480 sequences \\
        Sequence length & 1,024 tokens \\
        Total training tokens & 6.88B \\
        \bottomrule
    \end{tblr}
\end{table}

Training uses four NVIDIA A100 GPUs with distributed data parallelism and gradient accumulation. Each GPU processes microbatches of six sequences and accumulates gradients over 20 microbatches per optimizer update. We use BF16 mixed precision with cross-entropy computed in FP32.

\section{Data and Evaluation Protocol}
\label{app:data_evaluation}

\paragraph{Training data.}
The Pile~\citep{gao2020pile} combines 22 text sources covering web content, books, academic writing, code, and other domains. We use its publicly available \href{https://huggingface.co/datasets/EleutherAI/the_pile_deduplicated}{deduplicated version}, tokenized with the GPT-2 tokenizer and separated by end-of-text tokens. We reserve 2,636 complete documents, approximately 12.5M tokens, for internal validation and exclude training windows whose inputs or targets intersect these documents. SLT and RLT use the same training split containing 6.88B tokens, with identical training examples and data order.

\paragraph{Perplexity evaluation.}
For The Pile~\citep{gao2020pile}, we use its released test split. For FineWeb-Edu~\citep{penedo2024fineweb}, we use 8,192 nonempty documents from the first shard of the \texttt{sample-350BT} configuration. For OpenWebText~\citep{gokaslan2019openwebtext}, we use a validation set comprising 4,007 documents. Each evaluation corpus is tokenized and divided into 1,024-token windows.

To control evaluation cost across loop counts and experimental conditions, we use fixed 10\% evaluation subsets of the available token windows from each corpus, shared across all models and settings. Table~\ref{tab:rlt_perplexity_data} summarizes their sizes.

\begin{table}[htbp]
    \centering
    \caption{Corpora used for perplexity evaluation.}
    \label{tab:rlt_perplexity_data}
    \small
    \begin{tblr}{
        width = 0.80\linewidth,
        colspec = {X[l]Q[c]Q[c]Q[c]},
        row{1} = {font=\bfseries},
        colsep = 4pt,
        rowsep = 1pt,
        column{1} = {leftsep=10pt},
        column{Z} = {rightsep=10pt},
    }
        \toprule
        Corpus & Available windows & Evaluation windows & Evaluation tokens \\
        \midrule
        The Pile & 364,548 & 36,455 & 37.33M \\
        FineWeb-Edu & 8,437 & 844 & 0.864M \\
        OpenWebText & 4,331 & 433 & 0.443M \\
        \midrule
        Total & 377,316 & 37,732 & 38.64M \\
        \bottomrule
    \end{tblr}
\end{table}

Perplexity is computed as
\begin{equation}
    \mathrm{PPL}=\exp\!\left(
        -\frac{1}{N_{\mathrm{eval}}}
        \sum_{n=1}^{N_{\mathrm{eval}}}
        \log p_\theta(y_n\mid\mathcal C_n)
    \right),
    \label{eq:app_perplexity}
\end{equation}
where $N_{\mathrm{eval}}$ is the total number of evaluated target tokens, $y_n$ is a target token, and $\mathcal C_n$ is its preceding context within the corresponding evaluation window.

\paragraph{Downstream tasks and sample sizes.}
By default, our downstream evaluation uses a fixed subset of 1,866 examples across ten benchmarks, with ARC evaluated on its Easy and Challenge subsets. Table~\ref{tab:rlt_downstream_data} reports the original and evaluation sample counts. These subsets reduce evaluation cost while making sample counts more balanced across benchmarks. All models use the same selected examples within each experimental setting.

\begin{table}[htbp]
    \centering
    \caption{Downstream evaluation sample sizes and metrics.}
    \label{tab:rlt_downstream_data}
    \small
    \begin{tblr}{
        width = 0.80\linewidth,
        colspec = {Q[l]Q[c]Q[c]Q[l]},
        row{1} = {font=\bfseries},
        colsep = 15pt,
        rowsep = 1pt,
        column{1} = {leftsep=10pt},
        column{Z} = {rightsep=10pt},
    }
        \toprule
        Benchmark & Original size & Evaluation size & Metric \\
        \midrule
        COPA & 100 & 100 & Acc \\
        HellaSwag & 10,042 & 200 & $\mathrm{Acc}_{\mathrm{norm}}$ \\
        LAMBADA & 5,153 & 200 & Acc \\
        OpenBookQA & 500 & 200 & $\mathrm{Acc}_{\mathrm{norm}}$ \\
        PIQA & 1,838 & 184 & $\mathrm{Acc}_{\mathrm{norm}}$ \\
        RACE-high & 1,045 & 105 & Acc \\
        SciQ & 1,000 & 200 & $\mathrm{Acc}_{\mathrm{norm}}$ \\
        ARC-Easy & 2,376 & 238 & $\mathrm{Acc}_{\mathrm{norm}}$ \\
        ARC-Challenge & 1,172 & 117 & $\mathrm{Acc}_{\mathrm{norm}}$ \\
        Social IQA & 1,954 & 195 & Acc \\
        WinoGrande & 1,267 & 127 & Acc \\
        \bottomrule
    \end{tblr}
\end{table}

\paragraph{Prompt formats and scoring.}
Evaluation follows the task templates and scoring conventions of \texttt{lm-eval harness}~\citep{gao2023framework}. Multiple-choice tasks select the candidate with the highest conditional log-likelihood; normalized accuracy divides each candidate's score by its character length. LAMBADA uses the OpenAI version and requires the target word's tokens to match greedy predictions. SciQ prompts include the supporting passage, while RACE uses the high-school subset and includes preceding question--answer pairs from the same passage before scoring the final question. Contexts exceeding the model's input limit are truncated from the left.

We report the task-specific metric listed in Table~\ref{tab:rlt_downstream_data}. The ARC score is the arithmetic mean of the ARC-Easy and ARC-Challenge scores, and the overall downstream score is the unweighted mean across the ten benchmarks.

\section{Recurrent Stability Experiments}
\label{app:recurrent_stability}

All experiments use the frozen models obtained after 14,000 training steps. Unless otherwise specified, $K_{\mathrm{train}}=K_{\mathrm{test}}=8$, with the uniform time grid in Eq.~(\ref{eq:uniform_time_grid}). We write $K=K_{\mathrm{test}}$ below and define the RMS of a token representation $\mathbf{v}\in\mathbb{R}^{d}$ as
\begin{equation}
\mathrm{RMS}(\mathbf{v})=\sqrt{\frac{1}{d}\sum_{j=1}^{d}v_j^2}.
\label{eq:stability_rms}
\end{equation}

\subsection{Hidden-state Scale Measurement}
\label{app:stability_scale}

We measure hidden-state scales on 1,200 fixed windows of 1,024 tokens from the internal validation split of The Pile described in Appendix~\ref{app:data_evaluation}, yielding 1,228,800 token representations per loop. For each loop $i$, we extract the complete shared-block output $\mathbf{h}_i$ before feedback and the final RMS normalization. We compute two token-wise ratios, $\mathrm{RMS}(\mathbf{h}_i)/\mathrm{RMS}(\mathbf{h}_1)$ and $\mathrm{RMS}(\mathbf{h}_i)/\mathrm{RMS}(\mathbf{x}_0)$, using the corresponding first-loop output $\mathbf{h}_1$ or initial embedding input $\mathbf{x}_0$ from the same token and model as the reference. We then aggregate each ratio across tokens, reporting the mean, median, and 10th--90th percentiles at each loop.

\subsection{Perturbation Protocols}
\label{app:stability_noise}

We inject Gaussian noise with magnitude proportional to the RMS of the perturbed representation:
\begin{equation}
\mathcal{N}_{\rho}(\mathbf{v})=\mathbf{v}+\rho\,\mathrm{RMS}(\mathbf{v})\mathbf{z},\qquad \mathbf{z}\sim\mathcal{N}(0,\mathbf{I}_d),
\label{eq:stability_noise}
\end{equation}
where $\rho$ controls the noise level and scaling is applied separately to each token.

\paragraph{Initial-input perturbations.} We perturb the initial embedding input $\mathbf{x}_0$, which combines token and positional embeddings, before the first loop. We evaluate $\rho\in\{0.03,0.1,0.3,0.5,1,3,5,10\}$ on the three perplexity evaluation corpora described in Appendix~\ref{app:data_evaluation}. For each input, we compare the clean and perturbed final hidden outputs before the final RMS normalization, using the token-averaged Euclidean displacement
\begin{equation}
D_{\mathrm{out}}=\operatorname{mean}_{\mathrm{tokens}}\left\|\mathbf{h}_K^{\mathrm{noisy}}-\mathbf{h}_K^{\mathrm{clean}}\right\|_2.
\label{eq:stability_displacement}
\end{equation}

\paragraph{Recurrent perturbations.} We inject noise at every loop, either into the complete block output $\mathbf{h}_i$ before feedback or into the state $\mathbf{x}_i$ after feedback. Each injection uses fresh Gaussian noise scaled by the pre-injection RMS of the corresponding state on the current perturbed trajectory. Perturbations are also applied at the final loop, and the resulting vector is passed through the final normalization and language-model head for evaluation. We report changes in token-averaged cross-entropy and ten-task mean accuracy relative to each model's clean baseline, using the evaluation sets in Appendix~\ref{app:data_evaluation}.

Both perturbation experiments use three paired noise realizations across SLT and RLT. We first aggregate the evaluation results separately for each realization, then report their mean and sample standard deviation.

\subsection{Auxiliary MLP and Final-output Extrapolation for Contrastive Decoding}
\label{app:stability_loopcd}

\paragraph{Auxiliary prediction branch.} We train a separate auxiliary MLP for each frozen backbone and share its parameters across loops. Its input $\mathbf{u}_i$ is the output of the first physical Transformer layer at loop $i$, and its target $\mathbf{h}_i$ is the output of the complete three-layer block at the same loop. The branch is a bias-free SwiGLU MLP:
\begin{equation}
[\mathbf{a}_i,\mathbf{b}_i]=\mathbf{W}_{\mathrm{up}}\mathbf{u}_i,\qquad \hat{\mathbf{h}}_i=\mathbf{W}_{\mathrm{down}}\bigl(\operatorname{SiLU}(\mathbf{a}_i)\odot \mathbf{b}_i\bigr),\qquad \mathbf{a}_i,\mathbf{b}_i\in\mathbb{R}^{d},
\label{eq:stability_mlp}
\end{equation}
with $d=2{,}048$. The backbone remains frozen while the MLP is trained on its native trajectories using relative mean squared error:
\begin{equation}
\mathcal{L}_{\mathrm{MLP}}=\frac{1}{K}\sum_{i=1}^{K}\mathbb{E}_{\mathrm{valid\ tokens}}\left[\frac{\|\hat{\mathbf{h}}_i-\operatorname{sg}(\mathbf{h}_i)\|_2^2}{\|\operatorname{sg}(\mathbf{h}_i)\|_2^2+10^{-8}}\right],
\label{eq:stability_mlp_loss}
\end{equation}
where $\operatorname{sg}$ denotes stop-gradient.

The MLP is trained on the Pile training split for 2,000 steps, corresponding to approximately 0.98B tokens, with a global batch size of 480 and a sequence length of 1,024. We use AdamW with $\beta_1=0.9$, $\beta_2=0.95$, weight decay $0.01$, and gradient clipping at $1.0$. The learning rate increases to $3\times10^{-4}$ over 100 warmup steps and then follows cosine decay to $3\times10^{-5}$. We use the final MLP checkpoint and freeze both the backbone and the MLP during evaluation.

\paragraph{Final-output extrapolation for contrastive decoding.} After computing the unguided trajectory, we extrapolate the final output from either a native intermediate output or an MLP prediction:
\begin{equation}
\mathbf{h}_{\mathrm{ext}}=\mathbf{h}_K+\lambda(\mathbf{h}_K-\mathbf{r}_i),\qquad \mathbf{r}_i\in\{\mathbf{h}_i,\hat{\mathbf{h}}_i\}.
\label{eq:stability_loopcd}
\end{equation}
For RMS-aligned references, we replace $\mathbf{r}_i$ with
\begin{equation}
\mathcal{A}(\mathbf{r}_i;\mathbf{h}_K)=\mathbf{r}_i\frac{\mathrm{RMS}(\mathbf{h}_K)}{\max\{\mathrm{RMS}(\mathbf{r}_i),10^{-12}\}},
\label{eq:stability_alignment}
\end{equation}
computed separately for each token. We evaluate reference loops $i\in\{1,\ldots,8\}$ and strengths $\lambda\in\{0.1,0.2,0.3,0.4,0.5\}$, with $\lambda=0$ as the baseline. The extrapolated output is passed through the original final RMS normalization and language-model head. This intervention occurs only at the final readout and is not fed back into the recurrent computation.

Loop-wise contrastive decoding is evaluated on the ten downstream benchmarks. Prompt templates, task metrics, and the ten-task averaging procedure follow Appendix~\ref{app:data_evaluation}.

\subsection{Guidance during Recurrence}
\label{app:stability_guidance}

We use the auxiliary MLP from Appendix~\ref{app:stability_loopcd} to guide selected loops. At an activated loop $i$, both $\mathbf{h}_i$ and $\hat{\mathbf{h}}_i$ are computed from the current guided trajectory. We RMS-align the MLP prediction to the current block output and form
\begin{equation}
\mathbf{g}_i=\mathbf{h}_i+\lambda\bigl[\mathbf{h}_i-\mathcal{A}(\hat{\mathbf{h}}_i;\mathbf{h}_i)\bigr].
\label{eq:stability_guidance}
\end{equation}
At inactive loops, $\mathbf{g}_i=\mathbf{h}_i$. Following the feedback rules in Sections~\ref{subsec:looped_computation_and_state_feedback}--\ref{subsec:flow_inspired_state_feedback}, the next input is
\begin{equation}
\mathbf{x}_i=(1-\eta_{i-1})\mathbf{x}_{i-1}+\eta_{i-1}\mathbf{g}_i,\qquad \eta_{i-1}=\begin{cases}1,&\mathrm{SLT},\\[2pt]1/(K-i+1),&\mathrm{RLT}.\end{cases}
\label{eq:stability_guided_feedback}
\end{equation}

We evaluate guidance over loops $1$--$3$, $3$--$5$, $5$--$7$, and $1$--$7$. The final loop is unguided, and its output is decoded through the readout. Evaluation uses the downstream tasks described in Appendix~\ref{app:data_evaluation}.

\subsection{Quantization Configurations}
\label{app:stability_quantization}

We apply symmetric round-to-nearest quantization to the attention QKV and output projections and the two feed-forward linear layers per shared Transformer layer, covering twelve linear layers in total. Token and positional embeddings, the language-model head, conditioning modules, normalization layers, and feedback operations retain their original precision. Table~\ref{tab:stability_quantization} summarizes the configurations.

\begin{table}[htbp]
    \centering
    \caption{Quantization configurations for the shared Transformer blocks. Weight groups are formed along the input dimension within each output channel.}
    \label{tab:stability_quantization}
    \small
    \begin{tblr}{
        width = 0.90\linewidth,
        colspec = {Q[l]Q[l]Q[l]},
        row{1} = {font=\bfseries},
        colsep = 19pt,
        rowsep = 1pt,
        column{1} = {leftsep=10pt},
        column{Z} = {rightsep=10pt},
    }
        \toprule
        Configuration & Weight quantization & Activation format \\
        \midrule
        W8A16 & INT8, per output channel & BF16 \\
        W4A16 & INT4, groups of 128 & BF16 \\
        W8A8 & INT8, per output channel & Dynamic per-token INT8 \\
        \bottomrule
    \end{tblr}
\end{table}

Quantization scales are determined by the maximum absolute value within each channel or group, using integer ranges $[-127,127]$ for INT8 and $[-7,7]$ for INT4. W8A8 reuses the W8A16 quantized weights and dynamically quantizes the input to each target linear layer separately for each token. We simulate quantization through quantize--dequantize operations, with the dequantized tensors used in BF16 matrix multiplications. The same quantized weights are reused across loops and inference depths. Quantization is applied without calibration or retraining.

\clearpage

\section{Evaluation and Trajectory Analysis across Inference Depths}
\label{app:latent_dynamics}

\subsection{Inference-depth Evaluation}
\label{app:inference_depth_evaluation}

We evaluate the frozen SLT and RLT checkpoints trained with $K_{\mathrm{train}}=8$ at different inference depths. For each budget $K=K_{\mathrm{test}}$, we reconstruct the time grid with uniform step sizes according to Eq.~(\ref{eq:uniform_time_grid}) and update the step-size conditioning accordingly. Predictions are obtained from $\mathbf{h}_K$ through the final RMS normalization and shared language-model head.

\subsection{Trajectory Extraction and PCA Visualization}
\label{app:latent_pca}

\paragraph{Trajectory extraction.} We record recurrent trajectories on 16 fixed, nonoverlapping OpenWebText validation windows of 1,024 tokens, shared across models and budgets. Each trajectory contains the final token's full-block outputs $\mathbf{h}_1,\ldots,\mathbf{h}_K$ before feedback and final RMS normalization.

\paragraph{Shared projections across budgets.} For each model and window, we jointly fit a weighted PCA basis over an inference-budget set $\mathcal K_{\mathrm{fit}}$. Each budget receives equal total weight, with each hidden state weighted by
\begin{equation}
w_{K,i}=\frac{1}{|\mathcal K_{\mathrm{fit}}|K}.
\label{eq:latent_pca_weights}
\end{equation}
We center the pooled hidden states by their weighted mean and retain the first three principal components in $\mathbf{P}$. All budgets within a panel use this basis.

For visualization, we translate coordinates relative to the first-loop output at the matched depth:
\begin{equation}
\mathbf{z}_i^{(K)}=\mathbf{P}^\top\!\left(\mathbf{h}_i^{(K)}-\mathbf{h}_1^{(8)}\right).
\label{eq:latent_pca_projection}
\end{equation}
The matched-depth endpoint is $\mathbf{h}_8^{(8)}$.

\begin{figure}[!t]
\centering
\includegraphics[width=0.9\linewidth,height=0.91\textheight,keepaspectratio]{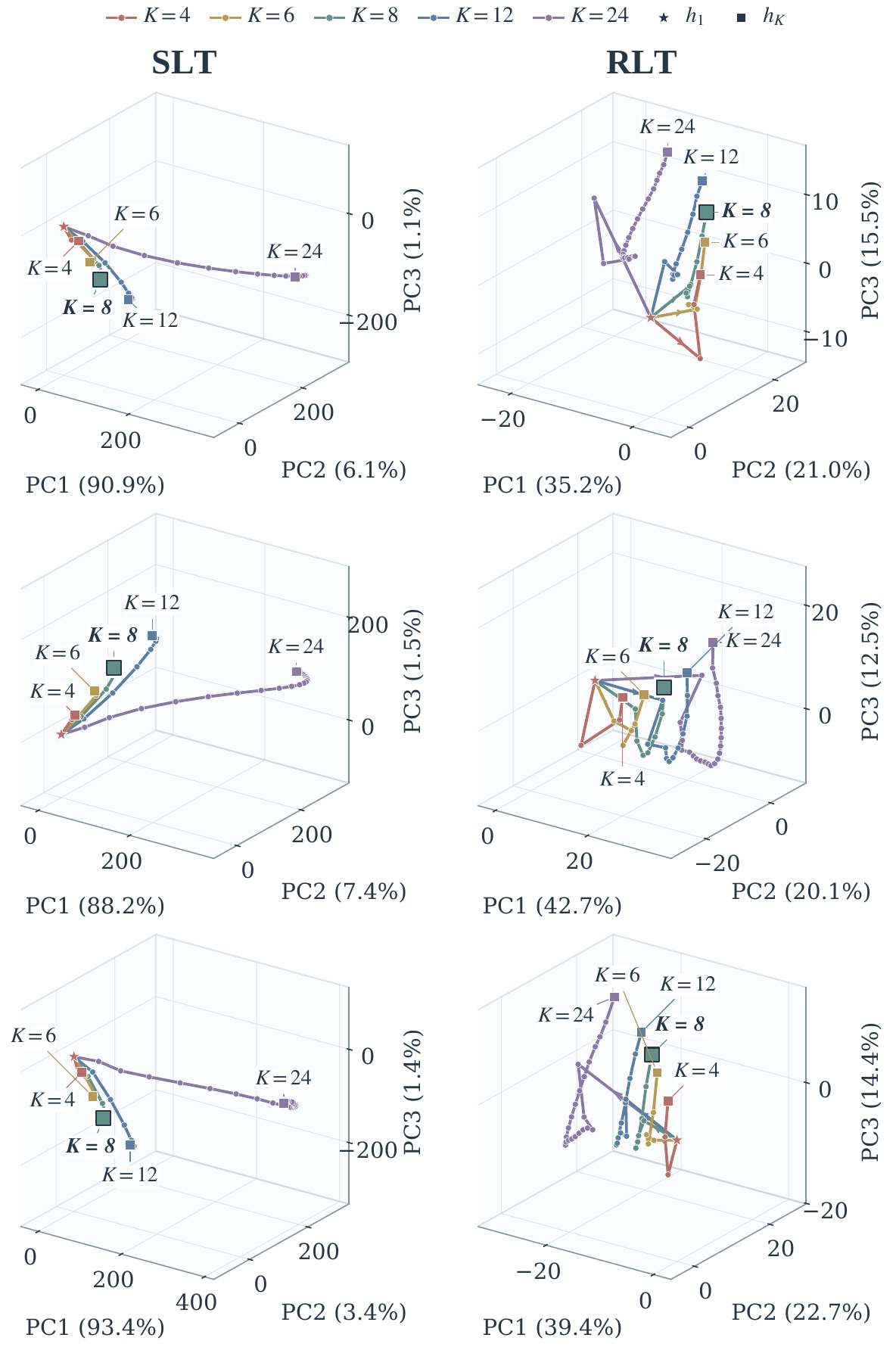}
\caption{3D PCA projections of final-token hidden outputs for three additional OpenWebText windows. Each row corresponds to one window; SLT and RLT occupy the left and right columns, respectively. Colors indicate the inference budget $K=K_{\mathrm{test}}$; stars and squares mark $\mathbf{h}_1$ and $\mathbf{h}_K$. The outlined green square marks the matched-depth endpoint.}
\label{fig:latent_pca_additional}
\end{figure}

\paragraph{Additional windows.} Figure~\ref{fig:latent_pca_additional} shows three additional OpenWebText windows using the same trajectory extraction and projection protocol as the main-text visualization. Across these examples, RLT dynamically adjusts its state-update step sizes, keeping its final hidden outputs close to the matched-depth reference as the inference budget changes. In SLT, smaller budgets stop short of the matched-depth endpoint, whereas larger budgets extend beyond it and substantially expand the trajectory along the dominant principal direction. These observations further support the claim that time-dependent state feedback enables RLT to adapt its recurrent updates to different inference budgets while maintaining output stability.

\end{document}